\documentclass[nohyperref]{article}

\usepackage{microtype}
\usepackage{graphicx}
\usepackage{booktabs} 
\usepackage{titlesec}

\usepackage{bibentry}
\usepackage{tikz}
\usetikzlibrary{shapes,arrows,positioning}
\usetikzlibrary{bayesnet}
\usepackage{hyperref}

\usepackage[accepted]{icml2023}

\usepackage{amsmath,amsfonts,amssymb,amsthm,bm}
\usepackage{mathtools}

\usepackage[capitalize,noabbrev]{cleveref}

\usepackage{caption}
\usepackage{subcaption}
\usepackage{comment}

\titlespacing{\paragraph}{0pt}{0pt}{0pt}

\usepackage{tikz}
\usetikzlibrary{trees,scopes,matrix,positioning}
\tikzset{
  mymx/.style={matrix of math nodes,nodes=myball,column sep=2.em,row sep=-1ex},
  myball/.style={draw,circle,inner sep=4pt},
  mylabel/.style={near start,sloped,fill=white,inner sep=1pt,outer sep=1pt,below,
    execute at begin node={$\scriptstyle},execute at end node={$}},
  plain/.style={draw=none,fill=none},
  sel/.append style={fill=green!10},
  prevsel/.append style={fill=red!10},
  route/.style={-latex,thick},
  selroute/.style={route,blue!50!green}
}

\usepackage{tikz}
\usetikzlibrary{shapes,decorations,arrows,calc,arrows.meta,fit,positioning}
\tikzset{
    -Latex, auto, node distance = 0.5 cm and 0.5 cm, semithick,
    state/.style = {circle, draw, minimum width = 0.6 cm},
    inter/.style = {rectangle, draw, minimum width = 0.7 cm, minimum height = 0.7 cm},
    point/.style = {circle, draw, inner sep = 0.04cm, fill, node contents = {}},
    bidirected/.style = {Latex-Latex,dashed},
    el/.style = {inner sep=2pt, align=left, sloped}
}

\RequirePackage{amsmath}
\RequirePackage{amssymb}
\RequirePackage{mathtools}
\ifx\proof\undefined
\RequirePackage{amsthm}
\fi
\RequirePackage{bm}
\RequirePackage{url}

\usetikzlibrary{backgrounds}

\usepackage{tikz}
\usetikzlibrary{trees,calc,fadings,decorations.pathreplacing, intersections}

\tikzset{%
  >=latex, 
  inner sep=0pt,%
  outer sep=2pt,%
  mark coordinate/.style={inner sep=0pt,outer sep=0pt,minimum size=3pt,
    fill=black,circle}%
}

\newcommand{\Ical}{\mathcal{I}}

\newcommand{\EE}{\mathbb{E}} 
\newcommand{\PP}{\mathbb{P}} 
\newcommand{\dimension}{{\bf{\rm dim}}}

\newtheorem{theorem}{Theorem}[section]
\newtheorem{proposition}{Proposition}[section]
\newtheorem{corollary}{Corollary}[section]
\newtheorem{lemma}{Lemma}[section]
\newtheorem*{remark}{Remark}
\theoremstyle{definition}
\newtheorem{definition}{Definition}[section]
\newtheorem{ass}[theorem]{Assumption}

\ifx\BlackBox\undefined
\newcommand{\BlackBox}{\rule{1.5ex}{1.5ex}}  
\fi

\ifx\QED\undefined
\def\QED{~\rule[-1pt]{5pt}{5pt}\par\medskip}
\fi

\ifx\proof\undefined
\newenvironment{proof}{\par\noindent{\bf Proof\ }}{\hfill\BlackBox\\[2mm]}
\fi
\ifx\proofsketch\undefined
\newenvironment{proofsketch}{\par\noindent{\bf Proof Sketch\ }}{\hfill\BlackBox\\[2mm]}
\fi

\theoremstyle{plain} 
\ifx\theorem\undefined
\newtheorem{theorem}{Theorem}
\numberwithin{theorem}{section}
\fi
\ifx\property\undefined
\newtheorem{property}{Property}
\fi
\ifx\corollary\undefined
\newtheorem{corollary}[theorem]{Corollary}
\fi
\ifx\lemma\undefined
\newtheorem{lemma}[theorem]{Lemma}
\fi
\ifx\proposition\undefined
\newtheorem{proposition}[theorem]{Proposition}
\fi
\ifx\assum\undefined

\fi
\ifx\definition\undefined
\newtheorem{definition}[theorem]{Definition}
\fi

\theoremstyle{remark} 
\ifx\remark\undefined
\newtheorem{remark}[theorem]{Remark}
\fi
\ifx\example\undefined
\newtheorem{example}[theorem]{Example}
\fi
\ifx\lemma\undefined
\newtheorem{lemma}[theorem]{Lemma}
\fi
\ifx\conjecture\undefined
\newtheorem{conjecture}[theorem]{Conjecture}
\fi
\ifx\claim\undefined
\newtheorem{claim}[theorem]{Claim}
\fi

\makeatletter
\renewcommand{\th@definition}{%
  \normalfont
  \thm@preskip 0.75\parskip \relax
  \thm@postskip 0pt \relax
}
\makeatother

\definecolor{rosso}{RGB}{220,57,18}
\definecolor{giallo}{RGB}{255,153,0}
\definecolor{blu}{RGB}{102,140,217}
\definecolor{verde}{RGB}{16,150,24}
\definecolor{viola}{RGB}{153,0,153}
\definecolor{babyblue}{RGB}{0,129,255}
\definecolor{darkgreen}{RGB}{6,148,60} 
\definecolor{darkblue}{rgb}{0.0,0.0,0.7} 

\hypersetup{
	colorlinks,
	breaklinks,
	linkcolor=red,
	urlcolor=darkblue,
	anchorcolor=darkblue,
	citecolor=darkgreen,
}
\providecommand*\url[1]{\href{#1}{#1}} 
\renewcommand*\url[1]{\href{#1}{\texttt{#1}}} 

\newcommand{\indep}{\perp \!\!\! \perp}

\newcommand{\instance}{\texttt{instance}}

\usepackage{todonotes}

\newcommand\rebuttal[1]{{\color{black}#1}}

\icmltitlerunning{On the Relationship Between Explanation and Prediction: A Causal View}

\begin{document}

\twocolumn[
\icmltitle{On the Relationship Between
           Explanation and Prediction: A Causal View}



\icmlsetsymbol{equal}{*}

\begin{icmlauthorlist}
\icmlauthor{Amir-Hossein Karimi}{mpi,eth,ggl}
\icmlauthor{Krikamol Muandet}{csp}
\icmlauthor{Simon Kornblith}{ggl}
\icmlauthor{Bernhard Sch\"olkopf}{mpi}
\icmlauthor{Been Kim}{ggl}
\end{icmlauthorlist}

\icmlaffiliation{mpi}{MPI for Intelligent Systems}
\icmlaffiliation{eth}{ETH Zurich}
\icmlaffiliation{ggl}{Google Research, Brain Team}
\icmlaffiliation{csp}{CISPA-Helmholtz Center for Information Security}

\icmlcorrespondingauthor{Amir-Hossein Karimi}{amir@tue.mpg.de}

\icmlkeywords{Machine Learning, ICML}

\vskip 0.3in
]



\printAffiliationsAndNotice{This work was primarily conducted when the first author was interning at Google Research.}

\begin{abstract}

Being able to provide explanations for a model's decision has become a central requirement for the development, deployment, and adoption of machine learning models. However, we are yet to understand what explanation methods can and cannot do. How do upstream factors such as data, model prediction, hyperparameters, 
and random initialization influence downstream explanations? While previous work raised concerns that explanations ($E$) may have little relationship with the prediction ($Y$), there is a lack of conclusive study to quantify this relationship. Our work borrows tools from causal inference to systematically assay this relationship.
More specifically, we study the relationship between $E$ and $Y$ by measuring the treatment effect when intervening on their causal ancestors, i.e., on hyperparameters and inputs used to generate saliency-based $E$s or $Y$s.
Our results suggest that the relationships between $E$ and $Y$ is far from ideal. In fact, the gap between `ideal' case only increase in higher-performing models--models that are likely to be deployed.
Our work is a promising first step towards providing a quantitative measure of the relationship between $E$ and $Y$, which could also inform the future development of methods for $E$ with a quantitative metric.
\end{abstract}

\section{Introduction and Related Work}
\label{sec:introduction}
Being able to provide explanations for a machine learning (ML) model's decision has become central to the development, deployment, and adoption of ML models. 
Explanations are important not only to help practitioners better understand the model's underlying rationale to debug models~\citep{adebayo2020,Rieger2020InterpretationsAU} and to influence the model's decision~\citep{koh2020concept,Bau2020,meng2022locating}, but also to ensure that models comply with regulatory requirements~\citep{gdpr16}.
However, Existing tools for interpretability have however elicited criticisms, often highlighting computational or qualitative user-study-based evidence that explanations generated from these tools may contain critical errors and must be used with care~\citep{poursabzi2018manipulating,chu2020visual,adebayo2018sanity,alqaraawi2020evaluating,srinivas2021rethinking, kindermans2019reliability}.

\begin{figure*}[t]
    \captionsetup[subfigure]{labelformat=empty} 
    \centering
    \begin{subfigure}[c]{.57\linewidth}
        \centering
        \resizebox{\linewidth}{!}{
        \begin{tikzpicture}[
            every node/.style={minimum width=2cm},
            every text node part/.style={align=center},
            polygon/.style= {shape=regular polygon, regular polygon sides=7, draw}
        ]
            \centering
            \node (X_10) [det] {archit.};
            \node (X_11) [det, below=of X_10, xshift=3mm, yshift=7mm] {$\ell_2$\\reg.};
            \node (X_12) [det, below right=of X_11, xshift=-5mm, yshift=8mm] {$\phi$};
            \node (X_13) [det, below right=of X_11, xshift=-10mm] {epochs};
            \node (X_14) [det, below left=of X_10, xshift=6mm, yshift=5mm] {drop-\\out};
            \node (X_15) [det, below right=of X_10, xshift=-5mm, yshift=6.5mm] {$w_0$\\$b_0$};
            \node (X_16) [det, above right=of X_10, xshift=-6mm, yshift=-6mm] {lr.};
            \node (X_17) [det, above left=of X_13, xshift=1mm, yshift=-10mm] {optmzr.};
            \node (X_19) [det, below left=of X_13, xshift=5mm, yshift=5mm] {data\\split};
            
            \draw [dashed,-] (1.75, -4.5) -- (1.75, 1);

            \node[draw] at (1.75, -5) {Training; $\mathtt{T}(\cdot)$};
            \node (X_4) [latent, minimum size = 2cm, right=of X_11, xshift=1mm] {model\\(trained)};
            \draw[solid] (X_10) edge[bend left=+23] node[yshift=0em] {} (X_4);
            \draw[solid] (X_11) edge[bend left=+5]   node[yshift=0em] {} (X_4);
            \draw[solid] (X_12) edge[bend left=0]   node[yshift=0em] {} (X_4);
            \draw[solid] (X_13) edge[bend left=-10] node[yshift=0em] {} (X_4);
            \draw[solid] (X_14) edge[bend left=+4] node[yshift=0em] {} (X_4);
            \draw[solid] (X_15) edge[bend left=+9] node[yshift=0em] {} (X_4);
            \draw[solid] (X_16) edge[bend left=+15] node[yshift=0em] {} (X_4);
            \draw[solid] (X_17) edge[bend left=-7] node[yshift=0em] {} (X_4);
            \draw[solid] (X_19) edge[bend left=-25] node[xshift=0mm] {} (X_4);
            \draw [dashed, -] (4.75, -4.5) -- (4.75, 1);

            \node[draw] at (4.75, -5) {Predicting; $\mathtt{P}(\cdot)$};
            \node (X_5) [det, below right=of X_13, xshift=10mm, yshift=8mm] {instance};
            \node (X_6) [latent, minimum size = 2cm, right=of X_4, xshift=1mm] {prediction};
            
            \draw[solid] (X_4) edge[bend left=0] node[yshift=0em] {} (X_6);
            \draw[solid] (X_5) edge[bend left=0] node[yshift=0em] {} (X_6);
            \draw [dashed,-] (8, -4.5) -- (8, 1);

            \node[draw] at (8, -5) {Explaining; $\mathtt{E}(\cdot)$};
            \node (X_7) [latent, minimum size = 2cm, right=of X_6, xshift=1mm] {explanation};
            \draw[solid] (X_4) edge[bend left=+37] node[yshift=0em] {} (X_7);
            \draw[solid] (X_5) edge[bend left=-15] node[yshift=0em] {} (X_7);
            \draw[solid] (X_6) edge[bend left=0] node[yshift=0em] {} (X_7);
        \end{tikzpicture}
        }
        \caption{}
        \vspace{-3mm}
        \label{fig:pipeline_detail}
    \end{subfigure}
    \hspace{5mm}
    \begin{subfigure}[c]{.30\linewidth}
                
        \begin{center}
            \begin{tikzpicture}
                \node[det, minimum size = 1.2cm] (h) at (0,0) {$H$};
                \node[det, minimum size = 1.2cm] (x) [below = 0.6 cm of h] {$X$};
                \node[state, minimum size = 1cm] (y) [below right = 0.2 cm and 0.8 cm of h] {$Y$};
                \node[state, minimum size = 1cm] (e) [right = 0.6 cm of y] {$E$};
                
                \path (h) edge (y);
                \path (x) edge (y);
                \path (y) edge (e);
                \path (h) edge [bend left  = 25] (e);
                \path (x) edge [bend right = 25] (e);
            \end{tikzpicture}
        \end{center}
    \caption{}
    \vspace{-3mm}
    \label{fig:pipeline_coarse}
    \end{subfigure}
    \vspace{-3mm}
    \caption{
    Explanation generating process involve three stages: training, predicting, and explaining (left). Intervening on factors ($H$, $X$) allow for studying their treatment effect (i.e., causal influence)on down-stream targets (i.e., $Y$, $E$) (right).
    }
    \label{fig:pipeline}
    \vspace{-5mm}
\end{figure*}
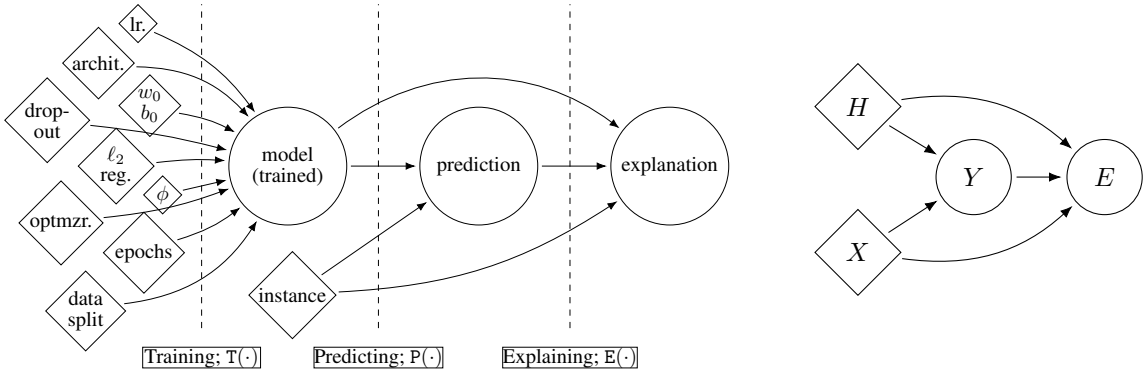

One focal point in many investigations is the relationship between explanations ($E$) and predictions ($Y$).
In this work, we seek to formalize this relationship, inspired by the common cause principle of \citet{Reichenbach1956} that states that if 
two variables are {\em statistically} dependent, there must be a common {\em cause} influencing both of them, and this common cause can be chosen such that it explains all the dependence. 
We develop a measure of dependence via the Potential Outcomes framework~\citep{rubin2005causal}. Viewed through a lens of causality, we evaluate the treatment effect of hyperparameters of the model, $H$ (i.e., $H$ taking on value $h'$, the counterfactual antecedent) on $E$ and $Y$ conditioned on a particular instance $x$. 
In other words, by measuring the treatment effect of each hyperparameter (e.g., choice of activation, initialization, training budget), we are measuring its influence on $E$ and $Y$, and in particular, how the influence is \textit{different or similar} in $E$ and $Y$ (Fig.~\ref{fig:pipeline}; left). 
Furthermore, under a careful evaluation, we tease apart the direct influence of $H$ on $E$ vs. its indirect influence mediated through $Y$ to better understand the flow of causation (Fig.~\ref{fig:pipeline}; right).

Why are hyperparameters considered treatments? Under a fixed random seed, hyperparameters are arguably the only reasonable causal ancestor of the model because they fully determine the weights of the resulting model and the behavior thereof.
They are also known to influence the inherent tendencies/performances of the model. 
For example, models trained on completely different hyperparameters could perform similarly under one metric (e.g., training loss), but have completely different task-specific performance, e.g., fairness~\citep{d2020underspecification}.
One can also use the hyperparameters alone to predict the final performance of the models~\citep{unterthiner2020predicting} or even use the model's weights to predict hyperparameters~\citep{Eilertsen2020}.

Our study reveals a surprising relationship between $E$ and $Y$ (precisely, measured by how a causal ancestor of the two influences them). 
In particular, for top-performing models, the influence on $E$ from $Y$ \textit{decreases} compared to relatively lower-performing models. For some methods, a causal ancestor of both $Y$ and $E$ directly influences $E$ much more than $Y$, leaving $Y$'s influence on $E$ minimal, even though this ancestor, i.e, hyperparameter, should not inform the explanation of the model in any way.
This finding was consistent across 30k pre-trained models with different hyperparameters across different datasets.
Our work informs practitioners on what different explanation methods can and cannot be used for: if one's goal is to find $E$ that is related to the prediction, $Y$, methods with little relationship between $E$ and $Y$ under our framework aren't the best choices. Our framework can also be used to drive the development of new methods by providing a quantitative metric.


\paragraph{Related Works}
Some studies argue that since many explanation methods claim to reveal a model's rationale behind its \textit{decision} ($Y$), there must be a ``strong correlation'' between $E$ and $Y$, e.g., when $Y$ changes significantly, $E$ must do so as well \cite{adebayo2018sanity,srinivas2021rethinking}, while others argue that $E$ should also reflect other factors in addition to $Y$ such as features in data points and data distribution~\citep{adebayo2018sanity, nie2018theoretical, srinivas2021rethinking, bilodeau2022impossibility}.
On the one hand, it has been observed empirically that explanations from an untrained model and a trained model can be visually and statistically indistinguishable~\citep{adebayo2018sanity}.
On the other hand, it was proven theoretically that $E$ has no relation to $Y$ in some cases  \citep{nie2018theoretical, srinivas2021rethinking}. However, quantitatively validating the relationship between $E$ and $Y$ while controlling for potential confounding factors such as hyperparameters and datasets remains an open question.

Despite some methodological similarities, our work is fundamentally different from using causal inference to \textit{generate} counterfactual explanations, e.g., \citet{wachter2017counterfactual},
where intervention is on the subset of features in an instance, rather than on a causal ancestor of $E$ while keeping the dataset constant. Our goal is to study the relation between $Y$ and $E$, and not to generate explanations.


\section{Methodology}
\label{sec:methodology}


To understand the relationship between $E$ and $Y$  via $H$'s impact on them, we perform an exploratory analysis on a class of ML models and then analyze their causal effects on the downstream $E$ and $Y$.

\vspace{-3mm}

\paragraph{Notation} Let $X\in\mathcal{X}\subseteq \mathbb{R}^d$ be a random variable representing a data instance and $H\in\mathcal{H}$ a random variable representing a hyperparameter vector. For $x\in\mathcal{X}$ and $h\in\mathcal{H}$, let $Y^*_h(x)$ and $E^*_h(x)$ be random variables representing respectively prediction and explanation associated with the hyperparameter value $h$ and data instance $x$. That is, 
$Y_h^*(x)$ and $E_h^*(x)$ correspond to the potential prediction and explanation when the model, trained with the hyperparameter vector $H=h$, is applied on the data point $X=x$. 
Put differently, the outcomes $Y^*_h(x)$ and $E_h^*(x)$ are realized by assigning the treatment (or intervention) $H=h$ (and the associated model) to the individual data $X=x$.
We distinguish $Y^*_h(x)$ and $E_h^*(x)$ from the notation of \emph{observed} prediction $Y_h(x) = Y^*_h(x)\,|\, H=h$ and explanation $E_h(x) = E^*_h(x)\,|\, H=h$ because in practice we cannot observe $Y^*_h(x)$ and $E_h^*(x)$ for all values of $h$.
The observed values of the prediction and explanation will be denoted by $\hat{y}_h(x)$ and $\hat{e}_h(x)$, respectively.


\subsection{Explanation Generating Process}

At a high level, the \emph{explanation generating process} (EGP) 
shown in \Cref{fig:pipeline} describes a mechanical system that is engineered to train an ML model given an initial set of hyperparameters, $h$, which yields a prediction $\hat{y}_h(x)$ and an explanation $\hat{e}_h(x)$ given a test instance $x$.
Formally, a supervised ML model is obtained through a \emph{training procedure} $\mathtt{T} : \mathcal{H} \times \mathcal{D} \rightarrow \mathcal{F}$ given a set of training hyperparameters and a dataset $\mathcal{D} := (\mathcal{X}, \mathcal{Y})$.
The training procedure typically contains initialization, optimization, and regularization.
%
Once trained, the model can predict the target of a given test instance $x$ via a \emph{prediction procedure} $\mathtt{P} : \mathcal{F} \times \mathcal{X} \rightarrow \mathcal{Y}$.
%
Finally, local explanations $e$ are the result of an \emph{explanation procedure} $\mathtt{E} : \mathcal{F} \times \mathcal{X} \times \mathcal{Y} \rightarrow \mathcal{E}$ applied to a tuple of a trained model, test instance, and predicted target, $\hat{y}_h(x)$.
%
Note the absence of noise variables; under a fixed random seed, the procedures above are deterministic.
%
%
%

Although these procedures may not be expressible in closed-form, e.g., one may not conclusively infer the trained weights of a neural network by only looking at the hyperparameters, each procedure is executable on a computer, e.g., the model weights can be obtained by training procedure under a training setting and given budget.
%



\subsection{Potential Outcomes Framework}
\label{sec:potential_outcomes_framework}

To study the causal effects of hyperparameters, we adopt the Potential Outcomes (PO) framework~\citep{rubin2005causal}.
Given the temporal precedence of hyperparameters over the trained model parameters and in turn over the prediction and explanation, one may alternatively view the mechanical system in \Cref{fig:pipeline_detail} as the causal system shown in \Cref{fig:pipeline_coarse} (with graphical and structural components).
In this framing, the \emph{causal influence} of up-stream factors (e.g., $H, X$) on down-stream targets (e.g., $Y,E$) can be measured as the \emph{treatment effect} of a factor (e.g., treatment $H= h$ vs. control $H = h'$), on the down-stream target.

In what follows, we will refer to $Y^*_h(x)$ and $E^*_h(x)$ as \emph{potential} prediction and explanation on an instance $x$ when the model is trained with the hyperparameter $h$. 
For any pair $h,h'\in\mathcal{H}$, the individual treatment effect (ITE), which quantifies the treatment effect of assigning two different parameters, can be defined as
\vspace{-2mm}
\begin{equation}
    \label{eq:ite}
    \text{ITE}_Y(x) = Y^*_h(x) - Y^*_{h'}(x)\text{.}
\end{equation}
Similarly defined, the treatment effect for explanation is denoted as ITE$_E$. 
In principle, it is possible to realize $Y_h^*(x)$ and $E_h^*(x)$ for all $h\in\mathcal{H}$ given unlimited computational resources. 
As a result, one can evaluate $\text{ITE}(x)$ in practice by contrasting the predictions of models trained on hyperparameters $h$ and $h'$.
However, when this process becomes computationally prohibitive, we might face the so-called \emph{fundamental problem of causal inference}, i.e., for each $x\in\mathcal{X}$, we can only observe $Y_h^*(x)$ and $E_h^*(x)$ for a small number of hyperparameters $h$, but not the other $h'\neq h$. 
Furthermore, we may not be able to interpret the observed differences between $Y$ and $E$ that arise from two different $H$ as a causal effect unless the assumption of \emph{ceteris paribus}, i.e., all else being equal, is fulfilled. Retraining almost identical neural networks with all possible values of hyperparameters is however computationally prohibitive. Instead, we perform an observational study on a model zoo, a large collection of pre-trained models \citep{unterthiner2020predicting, jiang2018predicting}, to study the relationship between $E$ and $Y$; see \Cref{sec:observational-study} for further discussion.

Since our research question seeks to investigate the impact of \emph{multiple, potentially-non-binary} treatments (e.g., set of numerical and categorical $H$) on the target prediction/explanation (see \Cref{fig:pipeline_detail}), we amend the treatment definitions above as follows:
\begin{align}
    \begin{split}
    & Y^*_{h=1}(x) - Y^*_{h=0}(x) \\
    & \qquad \parbox{5.5cm}{\footnotesize effect of $h = 1$ w.r.t $h=0$ on $x \in X$ \\ (single binary treatment)} \label{eq:extended_te_1} \\
    \end{split} \\
    \begin{split}
    & \EE_{m\not=n}\left[Y^*_{h=n}(x) - Y^*_{h=m}(x)\right] \\
    & \qquad \parbox{5.5cm}{\footnotesize effect of $h = n$ w.r.t $h\not=n$ on $x \in X$ \\ (single non-binary treatment)} \label{eq:extended_te_2} \\
    \end{split} \\
    \begin{split}
    & \EE_{h_{\setminus i}}\left[\EE_{m\not=n}\left[Y^*_{[h_i=n, h_{\setminus i}]}(x) - Y^*_{[h_i=m, h_{\setminus i}]}(x)\right]\right] \\
    & \qquad \parbox{5.5cm}{\footnotesize effect of $h_i = n$ w.r.t $h_i\not=n$ on $x \in X$ \\ (multiple non-binary treatments)} \label{eq:extended_te_3}
    \end{split}
\end{align}
\noindent which allows for answering queries of the form ``\emph{what is the treatment effect of optimizer choice $\nu_1$ as opposed to $\nu_2$ on the local prediction of $x$}?''.
Were the optimizer choice, $\nu$, to be the only hyperparameter in the system, this query would be answered by \eqref{eq:extended_te_2}.
In the setting of \Cref{fig:pipeline_detail}, however, \eqref{eq:extended_te_3} is employed to also marginalize out the effect of other $H$s.
Although these expressions average over multiple set of $H$s, they all refer to the prediction of the same individual (ITE); extensions to CATE and ATE, aggregated over $x~\sim~\mathcal{X}$, follow naturally.
To give \eqref{eq:extended_te_1}, \eqref{eq:extended_te_2}, \eqref{eq:extended_te_3} a causal interpretation, the following assumption is required.

\begin{ass}[Full exchangeability]\label{asp:exchangeability}
    $Y^*_h \indep H$ and $E^*_h \indep H$ for all $h\in \mathcal{H}$.
\end{ass}

For example, random assignment of $h$ \rebuttal{within a given range of values} $h$ makes $Y^*_h \indep H$ and $E^*_h \indep H$.
Although the treatment effects are identifable, evaluating them is computationally expensive.
To understand why, it helps to compare with with the setting of counterfactual explanations~\citep{wachter2017counterfactual}.
Whereas \citet{wachter2017counterfactual} contrast $Y^*_{h}(x)$ with $Y^*_{h}(x')$, which only requires the invocation of the \emph{predicting procedure} given a new \instance~(e.g., a forward pass through a neural network), our work instead contrasts $Y^*_{h}(x)$ with $Y^*_{h}(x')$, which invokes the \emph{training procedure} given a new $H$~setting (i.e., a full re-training).
In practice, computing power is limited and we may only have access to the predictions under a single model, say, $Y^*_{h}(x)$ and it can be prohibitively expensive to produce the prediction under a different model, $Y^*_{h'}(x)$, especially for large neural networks.

\rebuttal{
Note that the full exchangeability condition in \cref{asp:exchangeability} involves the ``counterfactual'' prediction $Y^*_h$ and explanation $E^*_h$ rather than the ``observed'' counterparts $Y_h$ and $E_h$.
The counterfactual variables $Y^*_h$ and $E^*_h$ describe the prediction and explanation one would observe had all instances in the entire population received the hyperparameters $h$ as a treatment.
Therefore, while in general, $Y_h(x)$ = $Y^*_h(x)$ and $E_h(x)$ = $E^*_h(x)$ can be random as well, e.g., if there is an exogenous noise, in our setting they are deterministic and randomness in the system arises only from the distribution of $X$ (sampled from some dataset).
As an analogy, imagine a treatment assigned to a patient: an individual outcome $Y^*_h(x)$ for each patient $x$ and the population outcome $Y^*_h$ can both be random, but the former (randomness in $Y^*_h(x)$) is missing in our setting.
}

\textbf{Kernelized treatment effect (KTE)}
\label{sec:kernelized_treatment_effect}
In addition to non-binary treatments, our work studies the effect of treatments on non-binary target variables ($Y^*_h(x)$ and $E^*_h(x)$) with dimensionality higher than that typically studied in the literature. 
%
For example, when $x$ is an image of size $d_1\times d_2$, $E^*_h(x) \in \mathbb{R}^{d_1\times d_2}$.
This means that \eqref{eq:extended_te_1} will yield a treatment effect \emph{vector} (or \emph{map}) as opposed to a \emph{scalar} treatment effect.
In order to compare the relative effect of hyperparameters in various settings, we extend the standard definitions of treatment effects once again, by replacing the subtraction operator in \eqref{eq:extended_te_1} with an alternative notion of dissimilarity between counterfactuals, i.e.,

\vspace{-7mm}
\begin{equation}
    \begin{aligned}
    \label{eq:kte}
        \left\|\phi(Y_h^*(x)) - \phi(Y^*_{h'}(x))\right\|^2_{\mathcal{G}} &= k(Y^*_h(x), Y^*_h(x)) \\
        &- 2k(Y^*_h(x), Y^*_{h'}(x)) \\
        &+ k(Y^*_{h'}(x), Y^*_{h'}(x)) \\
    \end{aligned}
\end{equation}
\vspace{-5mm}

%
%
where $\phi:\mathcal{Y}\to\mathcal{G}$ is the canonical feature map associated with a positive definite kernel $k:\mathcal{Y}\times\mathcal{Y}\to\mathbb{R}$, i.e., $k(y,y') = \langle \phi(y),\phi(y')\rangle_{\mathcal{G}}$ for $y,y'\in\mathcal{Y}$, and $\mathcal{G}$ is a reproducing kernel Hilbert space (RKHS) associated with the kernel $k$; see, e.g., \citet{scholkopf2002learning,Muandet20:CME,Park21:CoDiTE} for detailed exposition. 
Similar extensions can be applied to explanations as well as to \eqref{eq:extended_te_2} and \eqref{eq:extended_te_3} for multiple non-binary treatments.
In \Cref{sec:analysis}, we test various kernels $k$ to test the sensitivity of our analysis to the choice of kernel.
%
%
%
This enables us not only to work with the high-dimensional multivariate outcomes through positive definite kernels, but also to capture subtle effects of the hyperparameters on prediction and explanation that are beyond the mean effect. 
Applying  kernel is especially important when we compare $E^*_h(x)$, as comparing each spatially-related pixel value across different images is likely to not lead to a meaningful result.
%
Although \citet{zhao2021causal} propose an alternative approach that might be suitable for analyzing the causal effects of interest in our work (i.e., using partial dependency plots), they emphasize that it should not replace a random experiment or a carefully designed observational study.

\subsection{Observational Study}
\label{sec:observational-study}

In practice, we may not be able to compute $Y^*_h(x)$ and $E^*_h(x)$ for all $h\in\mathcal{H}$ because of the limit on computational resources. Hence, we face the fundamental problem of causal inference that prohibits us to exactly evaluate the ITE in \eqref{eq:ite}.
To this end, we will denote the \emph{observed} prediction and explanation by $Y_h(x) = Y^*_h(x)\,|\, H=h$ and $E_h(x) = E^*_h(x)\,|\, H=h$, respectively. Both \eqref{eq:extended_te_2} and \eqref{eq:extended_te_3} can be defined in terms of $Y_h(x)$ and $E_h(x)$, but the empirical estimates of these quantities may not correspond to the true treatment effects as Assumption \ref{asp:exchangeability} may not hold. We also state the common assumptions in the PO framework:
\begin{ass}[Unconfoundedness]\label{asp:unconfoundedness}
    There exists no unobserved confounder between $Y_h$ and $H$ (and $E_h$ and $H$).
\end{ass}

Since we will use a large collection of pre-trained models to assess the impact of hyperparameters on prediction and explanation, Assumption \ref{asp:unconfoundedness} guarantees that no unobserved common factors could have influenced the choice of hyperparameters and outcomes (i.e., prediction and explanation).

\textbf{Model zoos as data:}
In order to study the effect of hyperparameters on downstream $Y$ and $E$, one must first obtain a large collection of models which are the result of combinations of the hyperparameters under study.
Fortunately, such datasets already exist, namely, \emph{model zoo}s~\citep{unterthiner2020predicting, jiang2018predicting}.
We use the dataset provided by \citet{unterthiner2020predicting}, a large collection of existing models that have already been trained with pre-specified hyperparameters (see \cref{sec:model_zoo_details} for more detail).
%
%

\textbf{Direct vs. indirect influences\label{sec:direct_indirect}:}
As we can see from \Cref{fig:pipeline_coarse}, given the data instance $x$, there are two different paths from the hyperparameters $H$ to explanation $E$.
The former is a direct influence of $H$ on $E$, whereas the latter is an indirect influence mediated by the prediction $Y$.
To tell them apart, we propose the following simple analysis.
Let $(H_i(x),Y_i(x),E_i(x))_{i=1}^n$ be a collection of hyperparameters, corresponding predictions, and explanations, respectively. Then, we conduct the correlation analysis on this dataset, in particular, comparing the total influence of $H$ on $E$ vs. that of $H$ on $Y$ (\Cref{eq:extended_te_3}).
Next, we construct an artificial dataset by randomly permuting the predictions $Y_i(x)$ in the original data
%
and recomputing the corresponding explanations.
This gives us a new data set $(H_i(x), Y_{i}(x), \tilde{E}_{[i]}(x))$ where $\tilde{E}_{[i]}(x)$ is the recomputed explanation based on $Y_{[i]}(x)$, the permuted version of $Y_{i}(x)$.
Finally, we conduct the same correlation analysis on the permuted data set.
Because $Y_{[i]}(x)$ (random permutation of $Y_i(x)$) weakens the direct influence of $H_i(x)$ on $Y_i(x)$ as well as the direct influence of $Y_i(x)$ on $E_i(x)$,
careful comparisons between these correlations can reveal the extent to which the explanation $E$ relies on the prediction $Y$ (or on $H$); see \cref{sec:results} for further details.
Since the underlying relationships can potentially be non-linear, and we are comparing high-dimensional outcomes, i.e., $Y$ and $E$, in feature spaces, it is unclear how to adopt the classical mediation analysis~\cite{pearl2022direct}.
Our analysis only serves as an approximation thereof.
%

\begin{figure*}[h!]
    \centering
    \includegraphics[width=0.8\textwidth]{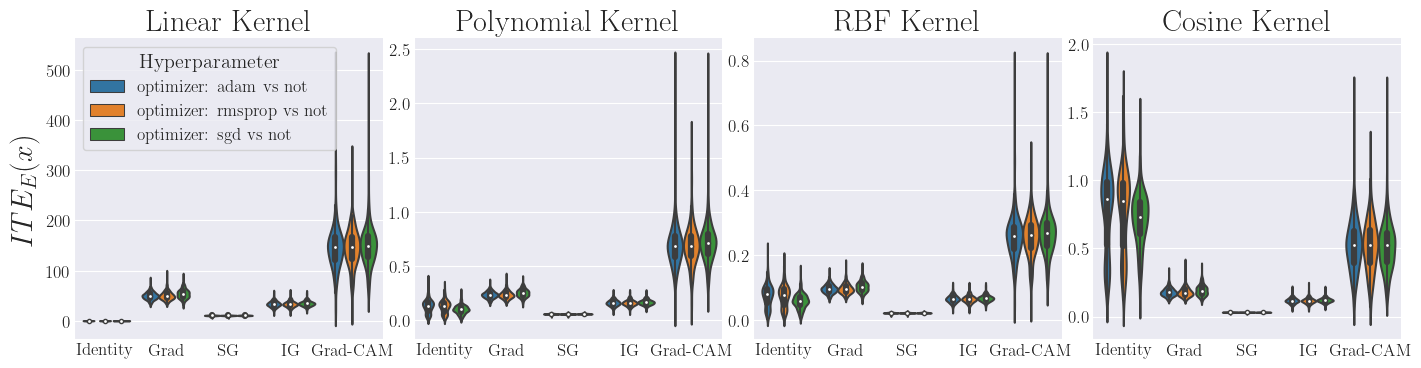}
    \vspace{-2mm}
    \caption{
        Comparison of the ITE$_E$ values with kernelized version of \eqref{eq:extended_te_3} obtained for $100$ instances from CIFAR10 for different choices of kernel (each column) shows that relative KTE values are not sensitive to the choice of kernels.
        %
        %
    }
    \label{fig:kernel_comparison}
    \vspace{-4mm}
\end{figure*}

\begin{figure*}[h!]
    \centering
    \includegraphics[width=\textwidth]{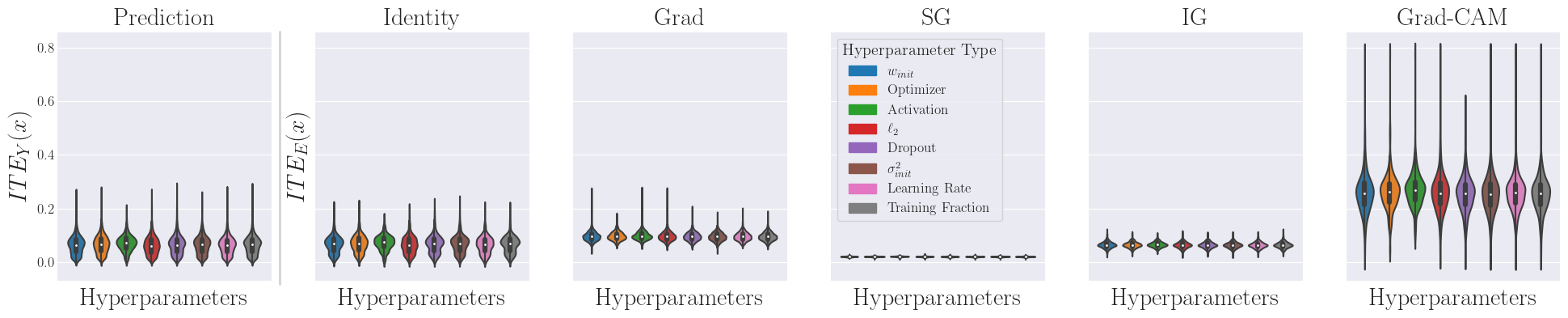}
    \vspace{-7mm}
    \caption{
        Comparison of ITE$_Y$ and ITE$_E$ for CIFAR10 shows that different types of $H$ influence $E$ and $Y$ in a similar way.
    }
    \vspace{-5mm}
    \label{fig:all_hparam_per_explanation_method}
\end{figure*}


\section{Analysis and Results}
\label{sec:analysis}

This section provides details of our analysis and results of our observational study in both global setting (all models) and local setting (models in each performance buckets). 

\subsection{Details of Observational Study}

\textbf{Model zoo dataset and pre-processing explanations}
\label{sec:model_zoo_details}
The dataset provided by \citet{unterthiner2020predicting} contains $30{,}000$ $3$-layer CNNs ($4{,}970$ parameters; weights and biases) that were trained until convergence (or a maximum of $86$ epochs) for multiple datasets.
%
The hyperparameters are drawn ``independently at random'' from pre-specified ranges.
Both the ranges and the training procedure are natural and resemble standard practice in machine learning, and the models are trained on commonly used CIFAR10, SVHN, MNIST, and FASHION MNIST datasets.
%
%
%
The random seed (for mini-batch GD sampling and for weight initialization) and the architecture of the base models are fixed throughout. 
The diversity of hyperparameters allows for a representative study of treatment effects (details in \cref{app:model_zoo}; \href{https://github.com/google-research/google-research/tree/master/invariant_explanations}{code}).
%

%
%
We study four commonly deployed saliency methods: 
\emph{gradient}~\citep{simonyan2013deep, erhan2009visualizing, baehrens2009explain}, 
\emph{SmoothGrad}~\citep{smilkov2017smoothgrad}, \emph{Integrated Gradients} (IG)~\citep{sundararajan2017axiomatic}, and \emph{Grad-CAM}~\citep{selvaraju2016grad}.
Note that many widely used methods are built based on these four methods~\cite{xu2020attribution, wang2021robust, simonyan2013deep}.
The generated explanation maps are preprocessed as in~\citet{adebayo2018sanity} (see \cref{app:model_zoo}). Since some methods only produce positive attributions, we zero out any negative attributions for the methods that produce both positive and negative values; this is so that we can compare all methods on an equal footing.
Finally, to measure the \emph{goodness} of treatment effect values, we introduce and evaluate a reference explanation method, namely \emph{Identity}, whereby $E$ is set to be identical to $Y$. Clearly, this is not a useful explanation for humans, but our goal here is to create an ideal $E$ that provides a point of comparison for our results. 


%




\subsection{Results}
\label{sec:results}


\textbf{KTE is not sensitive to the choice of kernel:}
KTE requires a decision on the type of kernel functions $k(\cdot, \cdot)$ (\Cref{sec:kernelized_treatment_effect}).
%
%
%
A natural question is whether in this context 
KTE is sensitive to the choice of kernel. We empirically compare the distribution of ITEs obtained (as per \eqref{eq:extended_te_3}) for 4 choices of kernels: (i) linear: $k(a,b) = a^T b$; (ii) polynomial: $k(a,b) = (\gamma a^T b + 1)^3$ (with $\gamma = 1/\dimension(a)$); (iii) RBF: $k(a,b) = \exp(-\gamma \|a - b\|^2)$; and (iv) cosine: $k(a,b) = a^T b / (\|a\|~\|b\|)$.
The results in \Cref{fig:kernel_comparison} suggest that the explanation ITE distributions are not sensitive to the choice of kernels that we tested.
%
Note that similar trends hold for other hyperparameters in \Cref{fig:all_hparam_per_explanation_method}.
We use the RBF kernel for the remainder of the paper.

\begin{figure*}[t]
    \centering
    \includegraphics[width=\textwidth]{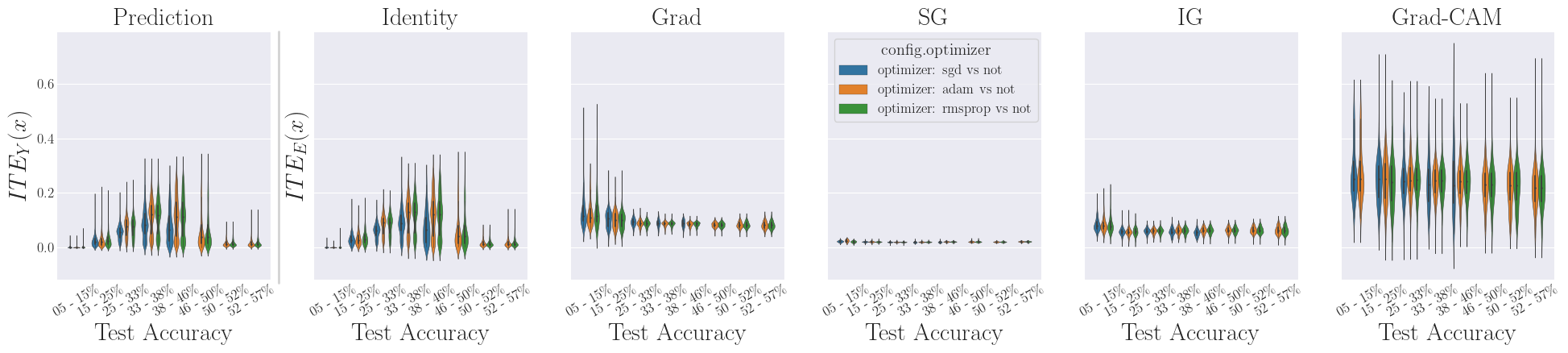}
    \includegraphics[width=\textwidth]{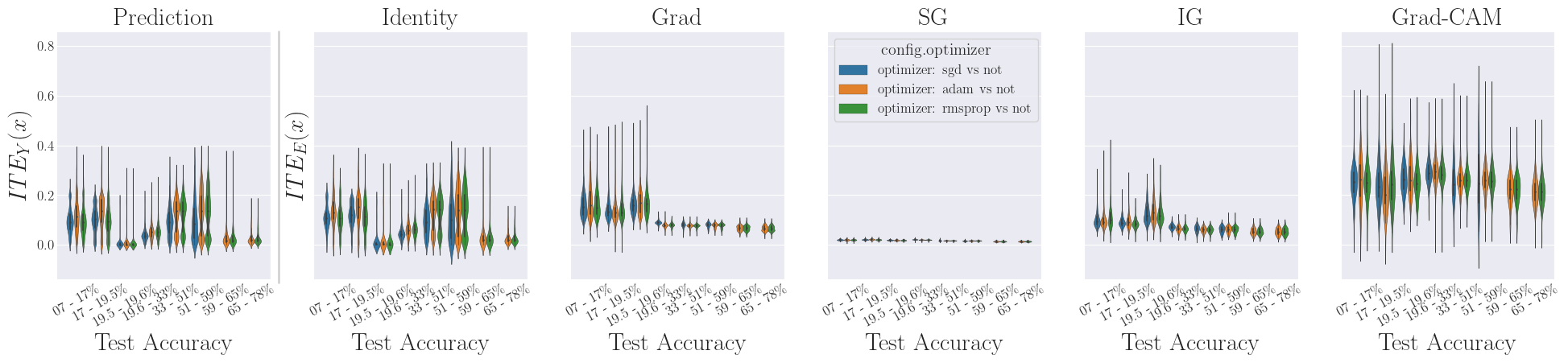}
    \vspace{-7mm}
    \caption{
        Comparison of ITE values of $h_\text{optimizer}$ on $Y$ (left) and $E$ (right) for models across different performance buckets, showing the discrepancy in the effect of $H$ on $Y$ vs. that on $E$ (top: CIFAR10; bottom: SVHN).
        \rebuttal{Interestingly, there is a difference of $\text{ITE}_\text{E}$ across accuracy buckets, and more importantly, none of the explainability methods resemble $\text{ITE}_\text{Y}$.} 
    }
    \label{fig:performance_comparison}
    \vspace{-5mm}
\end{figure*}

\textbf{Most types of $H$ influence $E$ and $Y$ in a similar way:}
Again, our goal is to measure the treatment effect of a causal ancestor ($H$) on $E$ and $Y$. 
The $H$ has different \textit{types} (e.g., initialization, activation, etc), and each type takes on multiple unique \emph{values} (i.e., treatment values) whose treatment effect on $Y$ or $E$ can be evaluated via \eqref{eq:extended_te_3}. As shown in \Cref{fig:all_hparam_per_explanation_method}, this effect is similar across different types of $H$ for both ITEs of $Y$ and $E$.
Stratifying the results per unique value of treatments also shows no apparent pattern, across all datasets considered (see \cref{fig_app:all_hparam_per_explanation_method}).

%

While this phenomenon may suggest that there is \textit{some} meaningful relationship between $E$ and $Y$, the pattern of $H$'s influence seems similar across different $H$. However, we notice that these `meaningful' relationships should only exist when $Y$ is meaningful (i.e., a trained network). In the next section, we divide these results into low/mid/high-performance buckets for further investigation.



\begin{figure*}[t]

    \captionsetup[subfigure]{labelformat=empty} 
    \centering
    \begin{subfigure}[b]{0.75\textwidth}
        \centering
        \includegraphics[width=\textwidth]{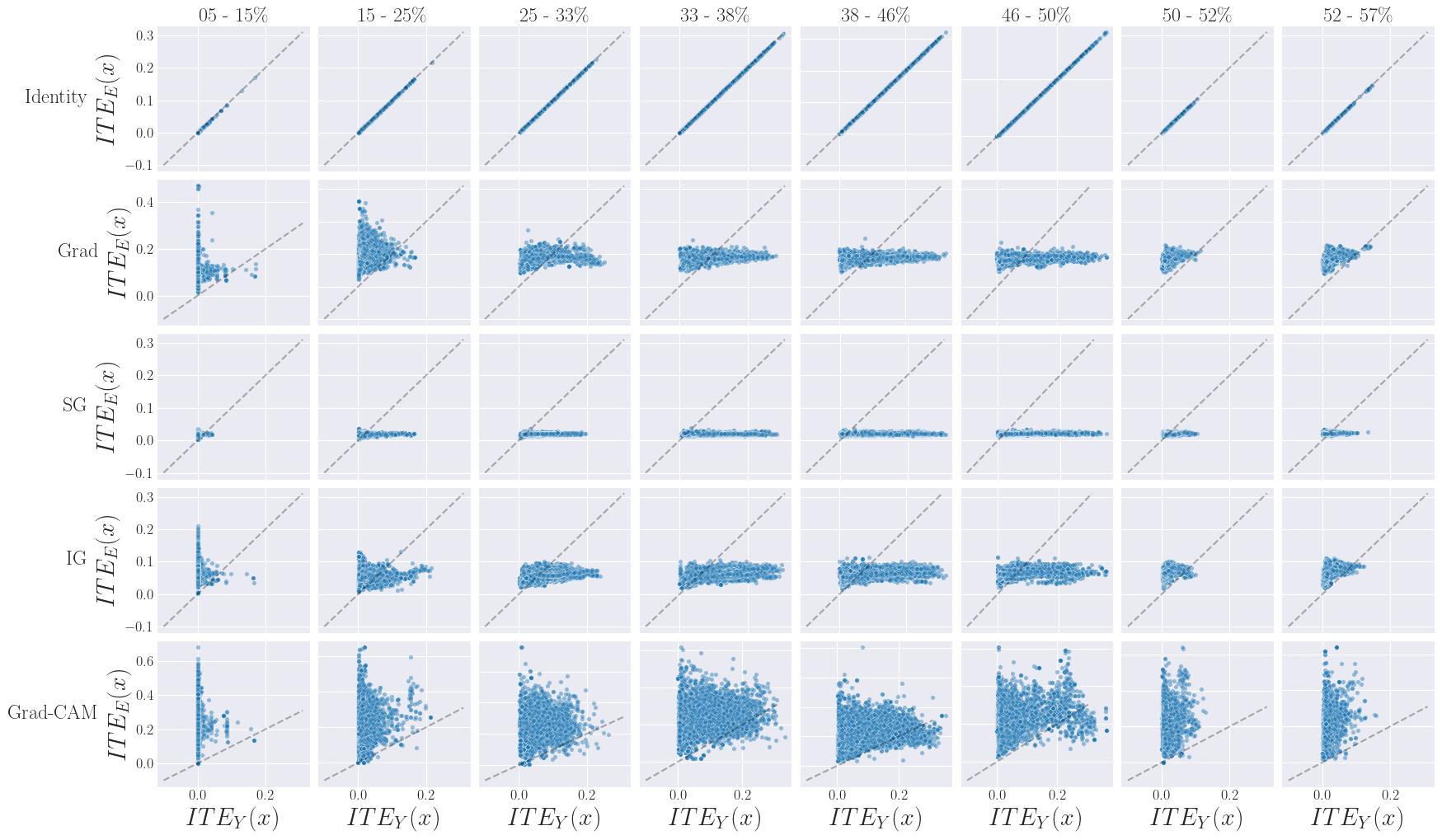}
        \caption{}
    \end{subfigure}
    \begin{subfigure}[b]{0.23\textwidth}
        \centering
        \includegraphics[width=\textwidth]{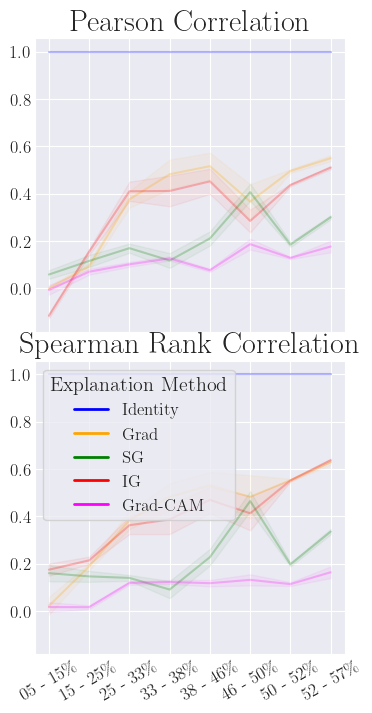}
        \caption{}
    \end{subfigure}

    \vspace{-7mm}
    \caption{
        (left) Each column is a subset of models at each accuracy bucket, each row is a different explanation method. 
        Whereas low-performing CIFAR10 models (first column) show little change in predictions as their explanations differ, top-performing models show the reverse of this trend.
        %
        (right) Correlation measures of the scatter plots on the left show a decreased correlation in the top 1\% models.
    }
    \label{fig:scatter_ite_y_vs_e}
    \vspace{-5mm}
\end{figure*}


\textbf{$H$ influences $Y$ (and $E$) differently across performance buckets:}
%
%
%
%
The relationship between $E$ and $Y$ when $Y$ is from an untrained model v.s. a trained model should be qualitatively different. 
Teasing out how much $Y$ influences $E$ is one of the long-standing questions in interpretability; some have argued that $E$ is visually indistinguishable when $Y$ is from trained or untrained models~\cite{adebayo2018sanity}. How the relationship between $E$ and $Y$ changes as a function of the performance of the model is important for practitioners in deciding when $E$ can or cannot be used. Thus, we conduct the remaining analysis by stratifying models into different accuracy buckets.
%
In particular, we stratified the $30{,}000$ models into 8 buckets according to their accuracies to observe the treatment effect in each group (\Cref{fig:performance_comparison}). We use
$0$-$20^\text{th}$, $20$-$40^\text{th}$, $40$-$60^\text{th}$, $60$-$80^\text{th}$ and $80$-$90^\text{th}$,
$90$-$95^\text{th}$, $95$-$99^\text{th}$ and $99$-$100^\text{th}$ percentiles as groups for all four datasets (finer granularity for top models that are more likely to be deployed; summarized in \Cref{tab:boundary_performance_buckets}).

\textit{The control group:}
%
Calculating ITE for each performance bucket requires a decision on control groups, i.e., the point of comparison. 
%
There are two natural choices 1)
select a control group within each accuracy bucket 
or 2) use the same control group across all buckets. 
Each choice means we are answering slightly different questions; (1) answers ``the effect of $h_i = n$ w.r.t. $h_i\not=n$ on $x \in X$ such that training on $h_i\not=n$ gives a similarly performing model'' while (2) answers ``the effect of $h_i = n$ w.r.t $h_i\not=n$ on $x \in X$ such that training on $h_i\not=n$ gives a model with baseline performance''. 
%
Although the latter enables comparison of performance buckets on similar footing, two factors are changing simultaneously: a) $h_i = n$ to $h_i\not=n$ and b) the change in performance bucket, 
making it difficult to tease apart hyperparameters' contributions to the ITE values.
Therefore, we continue with within-accuracy-bucket control groups, and refrain from comparing absolute values of ITE (for $Y$ or $E$) across buckets, but instead, look to \textit{relative} ITE values of $H$ on $Y$ and $E$ across buckets. 


As seen in \Cref{fig:performance_comparison}, while both ITE$_Y$s (first column) and ITE$_E$s (the remainder of columns) vary across 
accuracy buckets, they appear not to follow the same pattern.
%
%
%
This raises an important question: \emph{how does the relationship between $Y$ and $E$ (measured by treatment effect of $H$ on both) change as models' performance changes?}


\textbf{Understanding the (odd) relationship between ITE$_Y$ and ITE$_E$:}
%
%
We first investigate the extent of the relationship between $\text{ITE}_Y$ and $\text{ITE}_E$ by measuring their relative changes, before separating the direct influence of $H$ on $E$ from the indirect influence mediated through $Y$.

%
%
One way to compare $\text{ITE}_Y$ and $\text{ITE}_E$ is using scatterplots. \Cref{fig:scatter_ite_y_vs_e} (left) shows scatterplots for different performance buckets and explanation methods.
Since the absolute value of each ITE is not directly comparable (due to different domains for $Y$ and $E$, and different baseline control groups, as explained above), we summarize the scatter plot trends by measuring the Pearson and Spearman Rank correlations between the raw ITE values (\Cref{fig:scatter_ite_y_vs_e}; right).


We observe that compared to the case of the Identity method,\footnotemark~whereby there is a perfect correlation between $\text{ITE}_Y$ and $\text{ITE}_E$ (the diagonal $x=y$ line), no other method seems to remotely follow a similar pattern. For most of the methods, the range of $\text{ITE}_E$ values varies similarly regardless of low/mid/high accuracy models, while $\text{ITE}_Y$ naturally shrinks in high accuracy models, which can be explained by the models becoming similar in their predictions. 
The correlation coefficient tells a similar, but more concise, story. While the correlation increases for Grad and IG in the higher accuracy bucket, both show only moderate correlation compared to the reference point (Identity). It is also unclear how the relationship between $E$ and $Y$ is similar in mid-accuracy (e.g., 33\%) and top-accuracy models.
The pattern described above is shared across all types of hyperparameters across four datasets (see \cref{fig:scatter_ite_y_vs_e_app}
and \cref{fig:correlation_ite_y_vs_e_app}).

\footnotetext{We remind that while the Identity explanation is not useful for humans in any way, it helps us to understand what a ``good'' explanation (where $Y$ is a major factor in deciding $E$) may look like through the lens of the proposed ITE analysis.}

To summarize, the $\text{corr}(\text{ITE}_Y, \text{ITE}_E)$ increases as the model accuracy increases, suggesting that $E$ (for Grad and IG) becomes a better reflection of $Y$ in higher-performing models,\footnotemark~which is desired.
Despite this, the correlation values are substantially lower than a maximally informative explanation (i.e., the Identity method) suggests that \emph{explanations may still be explaining something other than the prediction}.

\footnotetext{At least in the manner in which \emph{changes in $E$} reflect \emph{changes in $Y$} as a result of \emph{changes in upstream $H$}}

\begin{figure*}[t]
    \centering
    \includegraphics[width=.8\textwidth]{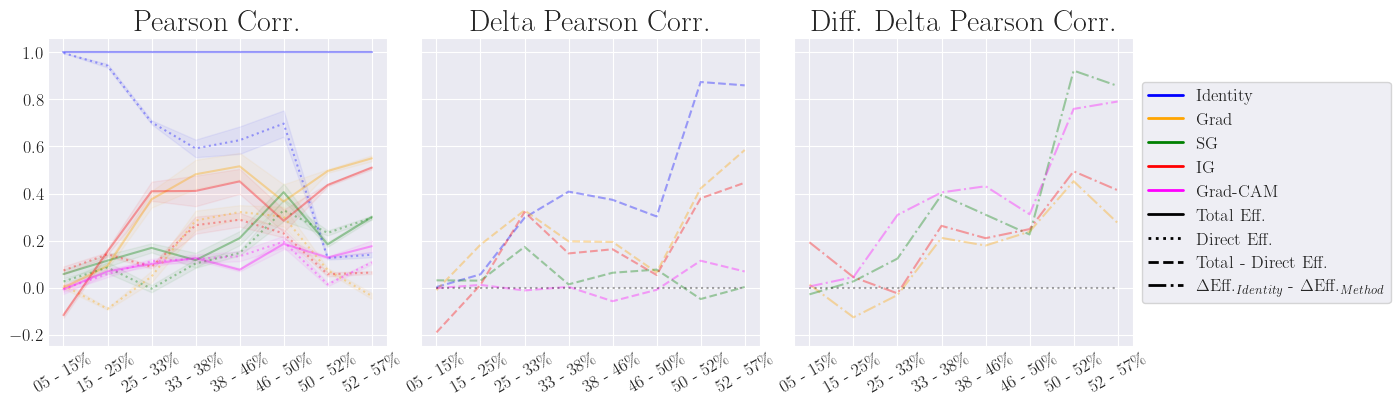}
    \vspace{-4mm}
    \caption{Pearson correlation between
    ITE$_Y$ and ITE$_E$ in total and direct effect (first column). 
    The second column is the difference between total and direct effect, where higher values 
    mean that the influence of $H$ on $E$ flows more through $Y$ (ideal). 
The third column plots the difference of delta correlations between ideal case (Identity) and each method. In other words, it indicates how far each method moves away from ideal case, as a model performs better. 
    }
    \label{fig:correlation_ite_y_vs_e}
   \vspace{-4mm}
\end{figure*}

\textbf{Direct vs. indirect influences:}
%
%
To understand how much of the explanation is reflecting the prediction, 
%
we can tease apart the effect of $H$ on $E$ that flows \emph{directly} vs. 
\emph{indirectly} through the prediction $Y$.\footnote{Since the individual for which $E$ is sought is fixed throughout (i.e., $X$ does not change; see discussion on identifiability at \Cref{app:identifiability}), we disregard the effect of $X$ on $E$ in this study.}
%
%
Intuitively, if explanations were only sensitive to $Y$, one would observe a \emph{low direct effect} and a \emph{high indirect effect}.
%
Conversely, a \emph{high direct effect} of $H$ on $E$ hints at the sensitivity of explanations to \emph{factors} not related to the prediction. 
Unlike all ITE$_E$ values we discussed so far that measures the \emph{total effect} of $H$ on $E$ (arising both directly and indirectly through $Y$), 
we ``sever'' the influence that $H$ has on $Y$ while retaining its effect on $E$.
As described in \Cref{sec:direct_indirect}, we compare $H$'s treatment effects on $E$ when $Y$ is and is not randomly permuted.

In the first column in \Cref{fig:correlation_ite_y_vs_e}, 
we first observe that none of 
explanations seem to follow the `ideal case' (Identity, 
$E$ is maximally informative of $Y$). 
The second column simply plots the difference between total and direct effects by subtracting direct effect from total effect (dotted line $-$ solid line in the first column).
This quantity roughly corresponds to the effect of $H$ on $E$ mediated through $Y$ (ideally, this value should be high in higher-performing buckets). 

What is even more concerning is \textit{how much} the difference between ideal case v.s., actual case \textit{worsens} in higher performing models. The third column plots this value: the difference between the ideal case (blue dotted line in the second column) and others. 
In other words, the higher a model performs, the more information for $E$ comes from something \textit{other than} $Y$. 
%
%
This is particularly concerning because these are models that are more likely to be deployed. For the case of SG and Grad-CAM, the influence of $H$ on $E$ mostly comes from $H$, not from the trained model or the prediction from it $Y$.
Putting it together, our comparison of direct and indirect influence reveals that the pattern of how $Y$ mediates the total influence of $H$ on $E$ is surprising and undesirable at times.

\section{Discussion and Conclusions}
\label{sec:discussion}

Our work investigates the relationship between $E$ and $Y$ using tools from causal inference.
In analyzing the treatment effect of a causal ancestor (i.e., $H$, determined prior to model training) of $E$ and $Y$ on them, the patterns observed for the direct and indirect influence reveals an undesirably high direct influence of $H$ on $E$ relative to influce of $Y$ on $E$. 
Our results suggest that the relationships between $E$ and $Y$ is far from ideal. In fact, the gap between `ideal' case only increases in higher-performing models--models that are likely to be deployed.
This means that there are \textit{other} factors that influence $E$ more than the prediction of the model, $Y$, and their influence becomes bigger and bigger as a model performs better.
If the users' goal is to understand the model's prediction, then most of the influence of $H$ on $E$ should be through $Y$ (note that \emph{which} $H$ should not influence $E$ is a decision by a user). The goal of our work is to first show that such influence exists in current models and present methods to perform quantitative analysis via the lens of the causal inference framework.

One can view our analysis as a more extensive, causal edition of~\citet{adebayo2018sanity}; we measure the treatment effect of $H$ on $E$ and $Y$ across $30{,}000$ models, while they quantitatively measure \textit{visual} similarities of $E$s as varying the quality of $Y$ in a single pair of models (trained and untrained). 
Furthermore, our analysis reveals that Grad-CAM (which arguably `passed' the sanity check in~\citet{adebayo2018sanity}) shows a worse correlation between the two ITEs across the buckets, meaning that the hyperparameters affect $Y$ and $E$ differently, hinting that no methods concretely outperform others.
Our results should be taken as a strong encouragement for practitioners to review other evidence instead of 
taking explanations at face value in their final decision-making.

\textbf{Limitations and Future Work}
%
%

The problem framing in \cref{fig:pipeline}, the formulations in \Cref{sec:methodology}, and the analytical framework presented over hyperparameter settings above naturally extend to any ML system (white-box or black-box) which have hyperparameters, $\mathcal{H}$, or more generally, any upstream \emph{factors}, that affect a final model.
%
The specific analyses presented in our paper, however, are bound by the choices made during the model zoo construction in \citet{unterthiner2020predicting}, e.g., choice and range/values of hyperparameters, and thus, the interpretation must be limited to the domain of $\mathcal{H}$ that we tested. 
For instance, while the model zoo offers an extensive number of models, their architecture is kept constant in all models (3 CNN layers, $\mathcal{O}(1e3)$ parameters).

Further studies on larger and complex models (e.g., \citep{frankle2018lottery, jiang2018predicting}) or similar analysis when the training dataset is (adversarially) changed (e.g., \citep{wang2021robust})
across different stages of training could reveal interesting insights.
%
%
%
%
Another valuable extension to our study is the analysis of our metric with other explainability metrics. We remark that our proposed metric assesses ``how much of the explanation is actually explaining the prediction,'' which, at least from an intuitive standpoint, is neither implied by nor implies other such metrics as \emph{intelligibility}, \emph{transparency}, \emph{complexity}, or \emph{user-friendliness}.
Finally, extending our work to  uncover the effect of hyperparameters~on other types of explanations would be interesting, e.g., influential samples~\citep{koh2017understanding}, Shapley values~\citep{lundberg2017unified}, concept-based methods~\citep{kim2018interpretability} surrogate-based methods, and recourse-based explanations and recommendations~\citep{karimi2020survey}. 
We believe the tools presented in our work may also be used to study the effect/influence of individual hyparparameters on model predictive performance prior to training.

%
%

\bibliography{refs.bib}
\bibliographystyle{icml2023}

\newpage
\appendix
\onecolumn
\label{app}
\section{Additional background material}
\label{app:additional_background}

\begin{figure}[t]
    \captionsetup[subfigure]{labelformat=empty} 
     \begin{center}
        \begin{tikzpicture}
            \node[det, minimum size = 1.1cm] (h) at (0,0) {$H$};
            \node[det, minimum size = 1.1cm] (d) [below = 0.6 cm of h] {$D$};
            \node[state, minimum size = 0.9cm] (x) [right = 0.6 cm of d] {$X$};
            \node[state, minimum size = 0.9cm] (w) [below right = 0.2 cm and 0.8 cm of h] {$W$};
            \node[state, minimum size = 0.9cm] (y) [right = 0.6 cm of w] {$Y$};
            \node[state, minimum size = 0.9cm] (e) [right = 0.6 cm of y] {$E$};
            
            \path (h) edge (w);
            \path (w) edge (y);
            \path (d) edge (x);
            \path (d) edge (w);
            \path (x) edge (y);
            \path (y) edge (e);
            \path (w) edge [bend left  = 35] (e);
            \path (x) edge [bend right = 25] (e);
        \end{tikzpicture}
    \end{center}
    \caption{
    Extended version of explanation generating process from \cref{fig:pipeline_coarse}, now with weights $W$ and dataset $D$ made explicit.
    }
    \label{fig:pipeline_appendix}
    \vspace{-3mm}
\end{figure}
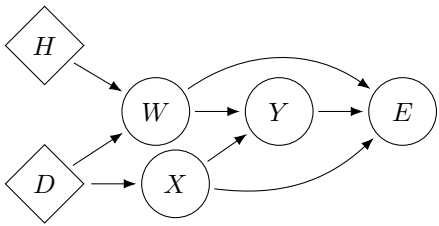

\subsection{The explanation generating process}

\rebuttal{To ease understandability, we refer to \cref{fig:pipeline_appendix} as the extended graph of \cref{fig:pipeline_coarse} which makes the weights $W$ and data $D$ explicit variables. Similar to \cref{fig:pipeline}, diamond nodes are considered factors whose effect we study, and circle nodes are random variables. 
In this extended graph, we clarify that $H$ is \emph{not} the model or trained weights. In other words, what we call hyperparameters ($H$) are sets like “\emph{method of optimization}: SGD or AdaGrad” or “\emph{regularizer coefficients}: 0.1 or 0.01 etc”. All $H$s can be assigned a value before we train any model and before observing any data.
Note that we do not have weights (denoted by $W$) in \cref{fig:pipeline_coarse}, as they are not the focus of our study; instead, we are interested in whether and how decisions made prior to training a model (i.e., assignments of $H$) influence downstream $Y$ and $E$.

Furthermore, considering the manner in which the model zoo was constructed whereby hyperparameters are sampled independently from some domain, there are no edges (no backdoors) from $X$ (or $D$) to $H$.
On the other hand, $W$ may be affected by the data distribution $D$, directly and/or through the training samples, but $W$ is not the focus of our work. Since we focus on the causal effect of hyperparameters $H$ on $Y$ and $E$ (not the weights $W$ on $Y$ and $E$), the formulations in \cref{sec:potential_outcomes_framework} remain unchanged.}


\subsection{On the identifiability and computability of treatment effects}
\label{app:identifiability}

An astute reader may notice that evaluating the treatment effects above as the difference between counterfactual contrasts bears a resemblance to another common explainability method, namely \emph{counterfactual explanations}~\citep{wachter2017counterfactual}.
This parallel is evident when thinking of \cref{fig:pipeline} in a coarser manner, i.e., $\mathcal{H},\mathcal{X} \rightarrow \mathcal{Y}$, whereby the hyperparameters~and dataset instance enter a \emph{potentially blackbox but queriable procedure} and yield a prediction.
Whereas the counterfactual explanations of \citet{wachter2017counterfactual} aim to identify minimal feature perturbations of the dataset instance under a fixed model (i.e., the hyperparameters~do not change; procedure: \emph{model prediction}), evaluating treatment effects as in \cref{eq:ite} is done by iterating over values of hyperparameters~to contrast resulting predictions given a fixed dataset instance (procedure: \emph{model training}).

Due to our mechanical setup, a number of interesting observations arise.
Although the \emph{training} ($\mathtt{T}$), \emph{predicting} ($\mathtt{P}$), and \emph{explaining} ($\mathtt{E}$) procedures may not be expressible in closed-form, the prediction $Y_h$ in \Cref{eq:ite} is exactly computable on a computer through \emph{forward simulation}.
In other words, upon selecting a set of hyperparameters, $H=h$, and under a fixed seed, all sources of randomness are controlled for and the procedures $\mathtt{T,P,E}$ deterministically yield a trained model, a prediction for a given instance, and the explanation for the said instance and model.
This is significant as it allows for the \emph{exact computation} of both $Y_{\textsc{treatment}}$ and $Y_{\textsc{control}}$ which is all that is needed to yield the value of the ITE exactly.
In other words, we can view both $Y_{\textsc{treatment}}$ and $Y_{\textsc{control}}$ as \emph{factual} outcomes.
Therefore, unlike real-world settings (e.g., taking a headache medication) where one cannot measure the ITE exactly (due to the impossibility of observing both \emph{factual} and \emph{counterfactual} outcomes simultaneously; whereby in such cases, the ITE is either approximated or the ATE is used instead.) the effect of all treatments, on both individual-level or population-level, are identifiable.

Although the treatment effects are \emph{identifable}, evaluating them is \emph{computationally expensive}.
To understand why, it helps to illustrate a parallel with the setting of counterfactual explanations~\citep{wachter2017counterfactual}.
Whereas the treatment effects in our setting (see \Cref{eq:ite}) contrasts $Y^*_{h}(x)$ and $Y^*_{h'}(x)$, the work of \citet{wachter2017counterfactual} contrasts $Y^*_{h}(x)$ and $Y^*_{h}(x')$.
Unlike the latter which only requires the invocation of the \emph{predicting procedure} given a new instance~$x$~(e.g., a forward pass through a neural network), the former invokes the \emph{training procedure} given a new hyperparameter~setting (i.e., a full re-training).
In practice, computing power is limited and we may only have access to the predictions under a single model, say, $Y^*_{h}(x)$ and it can be prohibitively expensive to produce the prediction under a different model, $Y^*_{h'}(x)$, especially for large neural networks.

\begin{table}[t!]
    \caption{Comparison of the classical and mechanical (our) setting for computing ITE values.}
    \begin{subtable}[t]{.47\textwidth}
        \centering
        \caption{In the classical setting for computing treatment effects, only one of the potential outcomes for each individual, $i$, is observable. The average treatment effect is defined as the average difference between individual treatment effects $ATE = \EE[Y_1^{(i)}] - \EE[Y_0^{(i)}]$.}
        \label{tab:classical_treatment_effect}
        \begin{tabular}{cccc}
            \hline
            $i$    & $Y_0$  & $Y_1$  & $Y_2$  \\ \hline
            1      & a      & -      & -      \\
            2      & -      & f      & -      \\
            3      & -      & -      & k      \\
            4      & -      & h      & -      \\
            \vdots & \vdots & \vdots & \vdots \\
        \end{tabular}
    \end{subtable}%
    \hfill
    \begin{subtable}[t]{.50\textwidth}
        \centering
        \caption{In our mechanical setting, given a model, $\hat{f}_h$, the potential outcome for any and all instances is computable (i.e., $\exists ~ Y_h(X_i), i \in \Ical \implies \exists ~ Y_h(X_k) ~ \forall ~ k \in \Ical$). Instead, one asks how to compute the treatment effect for $h'$ when no data is available for this hyperparameter.}
        \label{tab:proposed_treatment_effect}
        \begin{tabular}{cccc}
            \hline
            $i$    & $Y_0$  & $Y_1$  & $Y_2$  \\ \hline
            1      & a      & e      & -      \\
            2      & b      & f      & -      \\
            3      & c      & g      & -      \\
            4      & c      & h      & -      \\
            \vdots & \vdots & \vdots & \vdots \\
        \end{tabular}
    \end{subtable}
\end{table}

In order to reason about $Y^*_{h'}(x)$, one is compelled to instead ask a \emph{counterfactual} question: ``\emph{What would the prediction have been, had the optimizer been $\nu'$}?'' which can be answered through causal modeling without conducting real-world experiments, i.e., retraining with optimizer $\nu'$.
Metaphorically, there would have been no need for counterfactuals had one been able to simulate the entire universe (limited by either identification or computation).
It is the physical constraints that call for these counterfactuals.
Unfortunately, the procedures in \Cref{fig:pipeline} (left) are not available in closed form.
We clarify that unlike the classical randomized control trial (RCT) setting of evaluating ATE by contrasting average ITE values (where instances are randomly assigned to control or treatment), the mechanical nature of our setting allows for the target evaluation of all instances under control ($h$) or any treatment regime ($h'$); the challenge lies in the fact that applying a treatment to any one individual is as expensive as applying it to all individuals (see \Cref{tab:classical_treatment_effect} and \Cref{tab:proposed_treatment_effect} for comparison).
In this case, future research may explore the question of whether one can learn approximate procedures (i.e., approximate structural equations) to \emph{predict the predictions of an untrained classifier, given only its hyperparameters}.
In this regard, our preliminary results suggest a promising alternative to training individual models: developing meta-models that estimate a base model's prediction and explanation for an instance using only its hyperparameters, without actual training. This idea is derived from AutoML research, which predicts model accuracy based solely on hyperparameters, without training~\cite{unterthiner2020predicting}. As this issue rapidly evolves into a complex and multifaceted problem, we only briefly present the preliminary results here: a simple 3-layer MLP (namely, ``meta-model'') trained using $X$ and $H$ from a 10\% sample of models in the repository (i.e., 10\% of 30,000 ``base-models''), can estimate the predictions $Y$ for the rest of the base-models with an accuracy of approximately 45\%. It is important to note that the input features do not have trained weights and rely on hyperparameters instead, therefore saving compute. Furthermore, when the training is conducted on a subset comprising 10\% of the top-15\% performing models rather than on all models (with a mix of highly and poorly performing base models; refer to Table 2), the meta-model can predict the predictions $Y$ for the remaining base-models with an accuracy of around 80\%. Not only would this be a fascinating follow-up research project, but it would also hold substantial practical value for our framework.

\rebuttal{An implicit assumption made in \eqref{eq:extended_te_3} was that of mutual independence between hyperparameters, i.e., $h_i \indep h_j ~ \forall ~ j\not=i \implies h_{\setminus i} \sim \prod_{j\not=i} \PP(h_j)$.
This assumption yields an \emph{unconditional} treatment effect, whereby the causal effect of $h_i=\textsc{treatment}$ vs $h_i=\textsc{control}$ is averaged over all possible combinations of other hyperparameters, even if the combination rarely occurs in high-performing models.
In practice, however, it is conceivable that the hyperparameters~are selected carefully by the system designer and may be interpreted as being sampled from a distribution over hyperparameters, $\mathcal{H}$, internalized by the designer through \emph{prior} experience in training desirable models (e.g., accuracy, fairness).
%
%
Such down-stream criteria may act as a common child of the hyperparameters, inducing complex inter-dependencies (cf. Berkson’s paradox, \citep{pearl2009causality}).
In this case (i.e., $h_{\setminus i} \not\sim \prod_{j\not=i} \PP(h_j)$), the treatment effect answers such a query as ``among the set of hyperparameters~that yield models with at least $\gamma$ performance, what is the treatment effect of optimizer choice $\nu_1$ as opposed to $\nu_2$ on the local prediction of $x$?''
Therefore, whether or not we assume hyperparameters~to be mutually independent depends on the query being asked and assumptions made of the prediction/explanation generative process.
}
\rebuttal{Finally, one could consider straightforward extensions of \eqref{eq:extended_te_2} and \eqref{eq:extended_te_3} to support distributions over baseline control groups by adding an outer expectation that weights over the probability control group occurrence.}

\subsection{Model zoo details}
\label{app:model_zoo}

\rebuttal{For each of the 4 datasets (CIFAR10, SVHN, MNIST, FASHION) we consider $30{,}000$ pre-trained models, with diverse test accuracies resulting from the combinations of hyperparameters considered in the zoo \citep[Fig. 6]{unterthiner2020predicting}. We optionally analyze models stratified by their test performance, over 8 performance buckets; \Cref{tab:boundary_performance_buckets} shows the boundaries of these buckets.}

\begin{table}[t!]
    \centering
    \caption{\rebuttal{Test accuracy boundaries for each performance bucket for each dataset in the model zoo~\cite{unterthiner2020predicting}.}}
    \label{tab:boundary_performance_buckets}
    \begin{tabular}{ccccccccc}
        {\color{gray}percentile} & 0-20 & 20-40 & 40-60 & 60-80 & 80-90 & 90-95 & 95-99 & 99-100 \\
        \hline
        CIFAR10    & 5-15 & 15-25 & 25-33 & 33-38 & 38-46 & 46-50 & 50-52 & 50-57 \\
        SVHN       & 7-17 & 17-19.5 & 19.5-19.6 & 19.6-33 & 33-51 & 51-59 & 59-65 & 65-78 \\
        MNIST     & 4-11 & 11-35 & 35-73 & 73-89 & 89-95 & 95-96 & 96-97 & 97-98 \\ 
        FASHION   & 1-11 & 11-47 & 47-68 & 68-76 & 76-82 & 82-84 & 84-85 & 85-88 \\
    \end{tabular}
\end{table}
As a demonstration, \Cref{fig_app:ite_demonstration} shows the diversity in predictions of $30{,}000$ base models for a subset of CIFAR10 images for $1$ randomly sampled datapoint from each class.
It is noteworthy that the non-kernelized ITE values of \eqref{eq:extended_te_3} can be read directly from the figure, by contrasting the mean (shown in diamond) of each pair of nested bar plots (via application of linearity of expectations to \eqref{eq:extended_te_3}).

\textbf{Pre-processing explanations and other details}
To study the effect of hyperparameters on explanations,
we generate explanations, $E_h(x)$, via saliency-based methods.
In particular, the Gradient~\citep{simonyan2013deep, erhan2009visualizing, baehrens2009explain} and its smooth counterpart, SmoothGrad~\citep{smilkov2017smoothgrad}, Integrated Gradient (IG)~\citep{sundararajan2017axiomatic}, and Grad-CAM~\citep{selvaraju2016grad} methods are used due to their commonplace deployment\footnote{All methods are openly accessible here: \url{https://github.com/PAIR-code/saliency}.}~\citep{adebayo2018sanity}. Note that many other widely used methods are based on these four methods~\cite{kapishnikov2021guided, xu2020attribution, wang2021robust, simonyan2013deep}.
The generated explanation maps $E_h(x)$ are then processed to first remove outliers (via percentile clipping the values above 99th percentile), following by normalizing all attributions to fall in $[0,1]$. For Grad-CAM which only generates positive attributes, this is straightforward; for other methods that give positive and negative attributes (as each carry different semantics; contributing towards/against the prediciton), we first normalize to $[-1,1]$ and then clip any value below $0$.

The set of hyperparameters considered include the choice of optimizer, $w_0$  type, $w_0$ std., $b_0$  type, choice of activation function, learning rate, $\ell_2$ regularization, dropout strength, and dataset split~\citep[see][Appendix A.2]{unterthiner2020predicting}. To evaluate treatment effects as per \eqref{eq:extended_te_3}, continuous features are discretized by (log-)rounding to the nearest predetermined marker from within the range of the feature.\footnote{The following markers are used for (log-)rounding continuous features: $\ell_2$ reg.: $[1e^{-8}, 1e^{-6}, 1e^{-4}, 1e^{-2}]$, dropout: $[0, 0.2, 0.45, 0.7]$, $w_0$ std.: $[1e^{-3}, 1e^{-2}, 1e^{-1}, 0.5]$, learning rate: $[5e^{-4}, 5e^{-3}, 5e^{-2}]$.}

\textbf{Relation to other explainability metrics}

There are many such heuristics for rating explainability, and we recognize the absence of such comparisons in our research study.
At the same time, we emphasize that our proposed metric assesses ``how much of the explanation is actually explaining the prediction,'' which, at least from an intuitive standpoint, is neither implied by nor implies other such metrics as \emph{intelligibility}, \emph{transparency}, \emph{complexity}, or \emph{user-friendliness}. 
We also recognize that relying solely on the suggested metric may lead to misleading results and should not be considered adequate for endorsing an explanation approach. As demonstrated in footnote 1, we provide an instance where the Identity explanation implies an ideal correlation between $ITE_E$ and $ITE_Y$, even though it does not offer a meaningful explanation.
We encourage further investigation in this direction for future research.

\section{Additional experimental results}
\label{app:additional_experiments}

In this section, we present additional experimental results to complement those in the main body across different data dimensions or on new datasets.

\begin{figure}[h]
    \captionsetup[subfigure]{labelformat=empty} 
    \centering
    \begin{subfigure}[b]{0.75\textwidth}
        \centering
        \includegraphics[height=3.05cm]{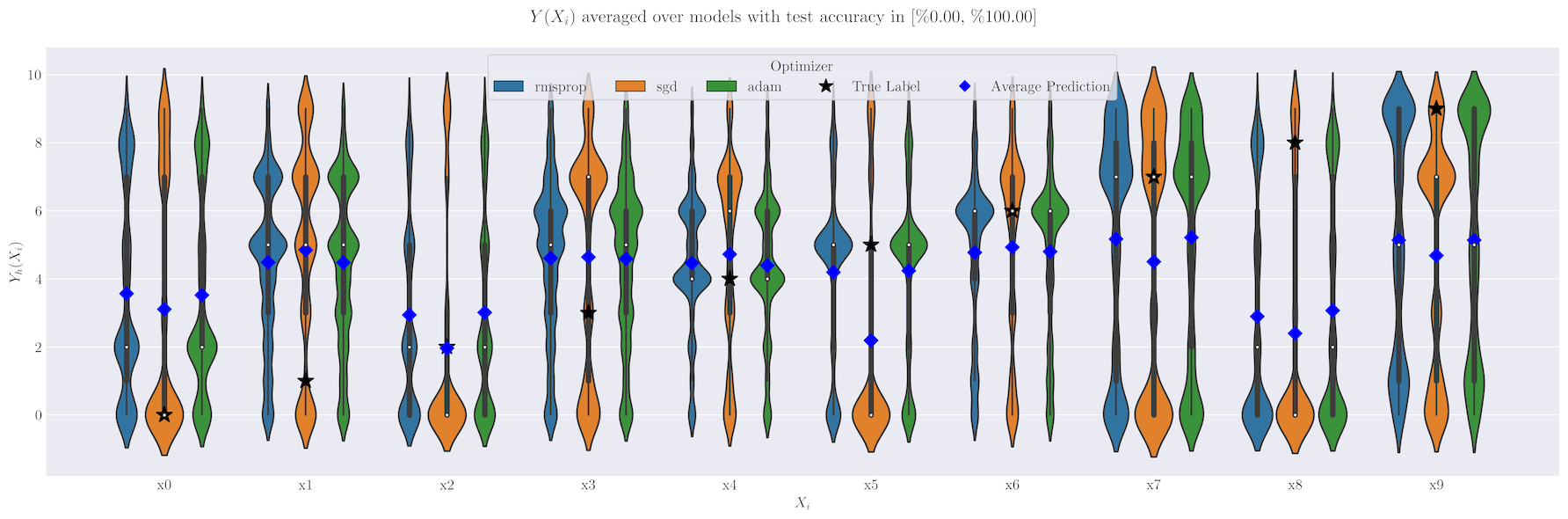}
        \includegraphics[height=3.05cm]{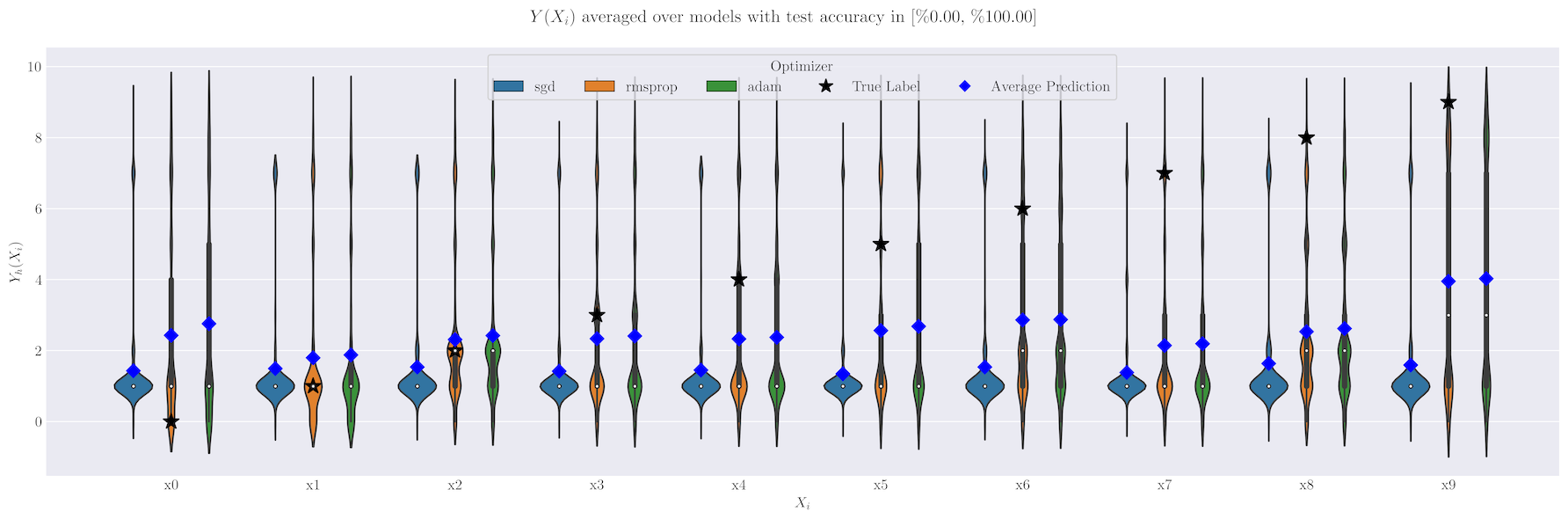}
        \includegraphics[height=3.05cm]{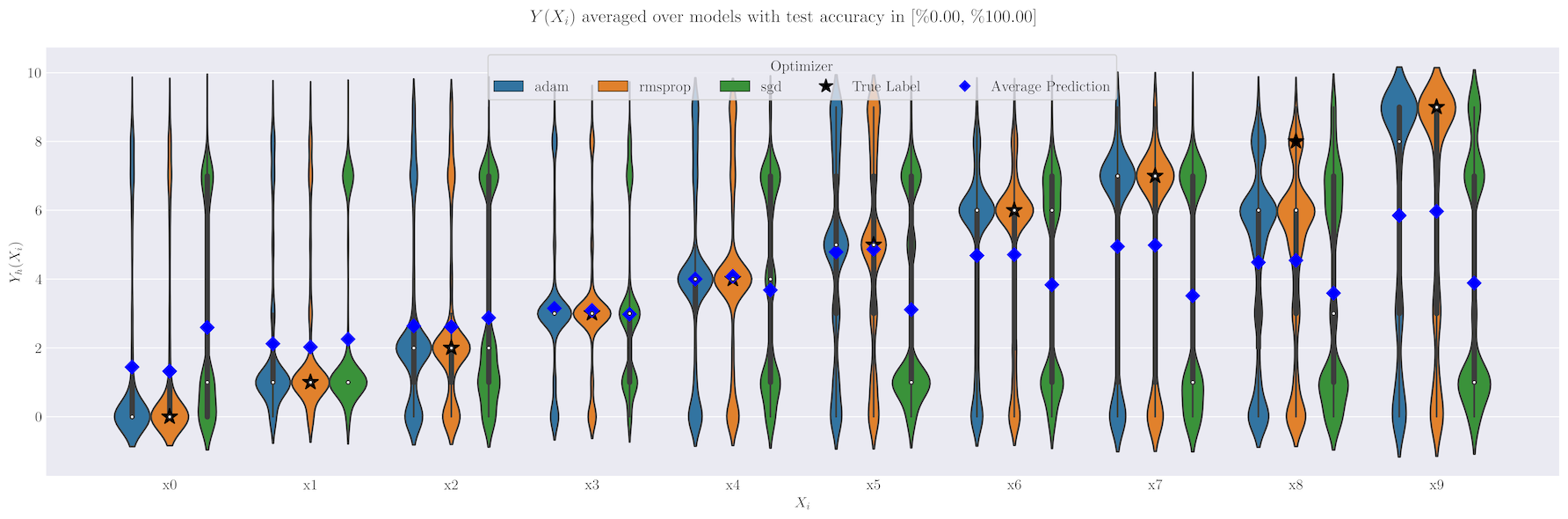}
        \includegraphics[height=3.05cm]{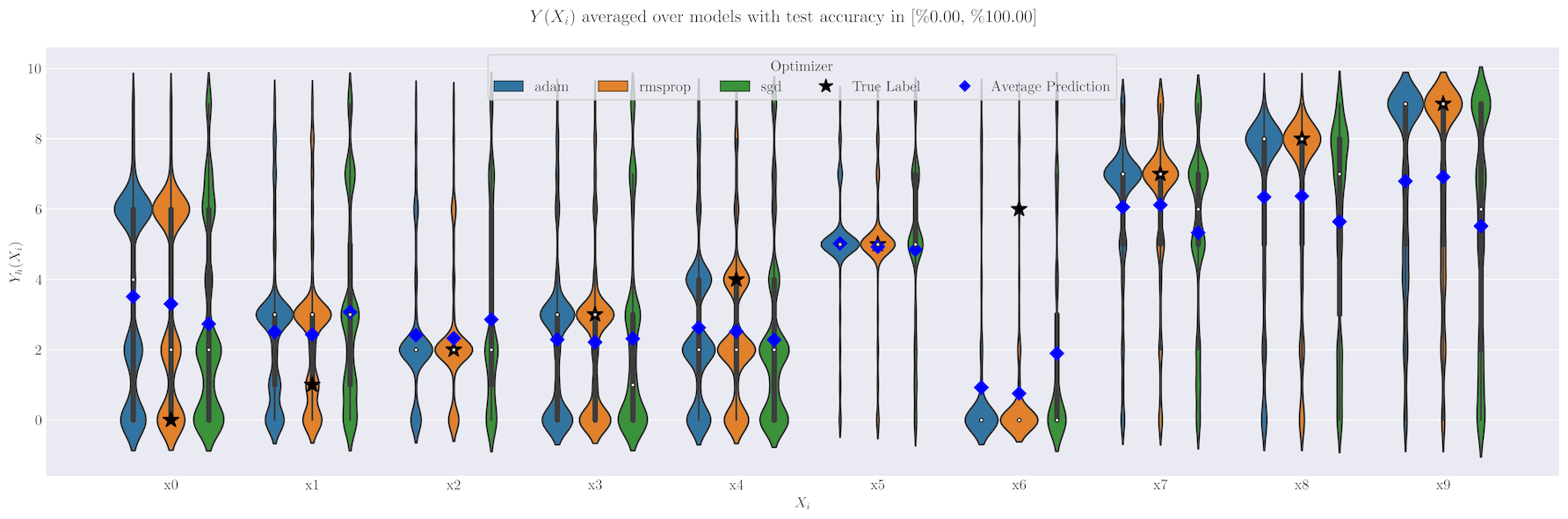}
        \caption{}
        \label{fig_app:performance_comparison_y_pred}
    \end{subfigure}
    \hfill
    \begin{subfigure}[b]{0.24\textwidth}
        \includegraphics[height=3cm]{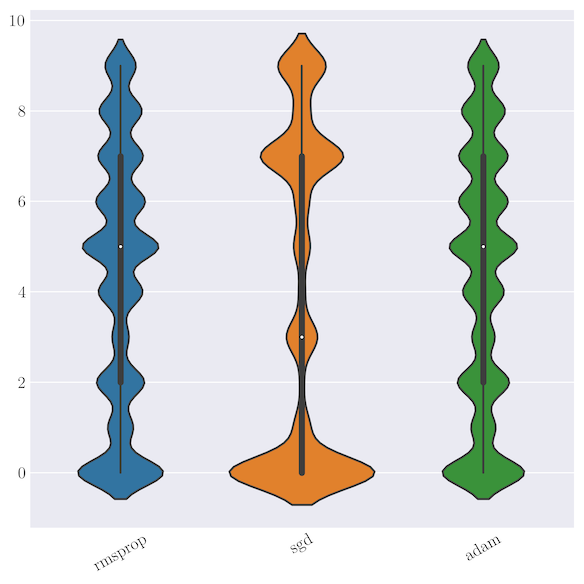}
        \includegraphics[height=3cm]{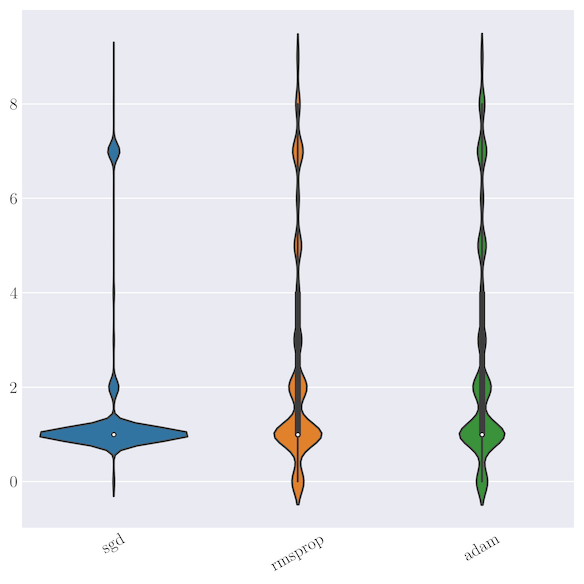}
        \includegraphics[height=3cm]{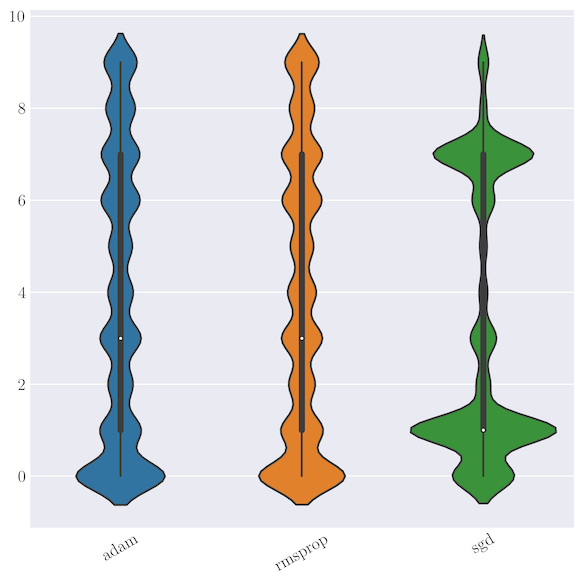}
        \includegraphics[height=3cm]{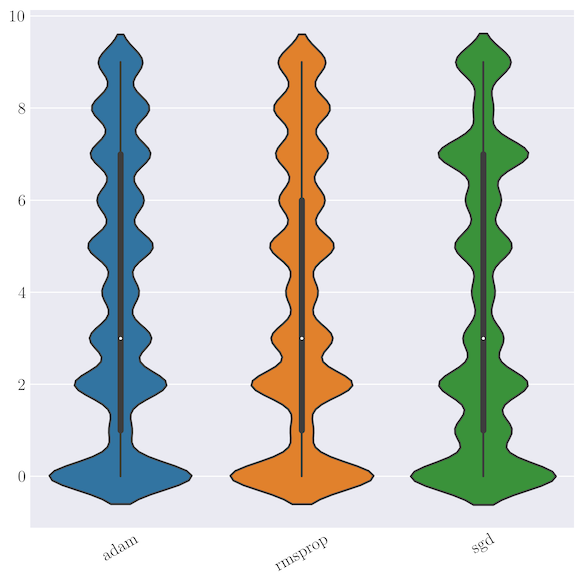}
        \caption{}
        \label{fig_app:performance_comparison_explan}
    \end{subfigure}
    \caption{
        The distribution of $Y_{h}(x_i)$ for a subset of $10$ random instances(1 per class) on $30{,}000$ base models (row 1: CIFAR10; row 2: SVHN; \rebuttal{row 3: MNIST; row 4: FASHION}).
        For each instance, each column holds the value of $h_\text{optimizer}$ fixed at one of $m$ unique values pertaining to this hyperparameter, while unconditionally iterating over other hyperparameters.
        In this manner, the difference in predictions across values of the hyperparameter, both at an individual (left) and aggregate level (right) can be attribute to, and only to, changes in this hyperparameter.
    }
    \label{fig_app:ite_demonstration}
\end{figure}

As a demonstration, \Cref{fig_app:ite_demonstration} shows the diversity in predictions of $30{,}000$ base models for a subset of CIFAR10 (top) and SVHN (bottom) images for $1$ randomly sampled datapoint from each class.
It is noteworthy that the non-kernelized ITE values of \eqref{eq:extended_te_3} can be read directly from the figure, by contrasting the mean (shown in diamond) of each pair of nested bar plots (via application of linearity of expectations to \eqref{eq:extended_te_3}).

\begin{figure}[h]
    \centering
    \includegraphics[width=\textwidth]{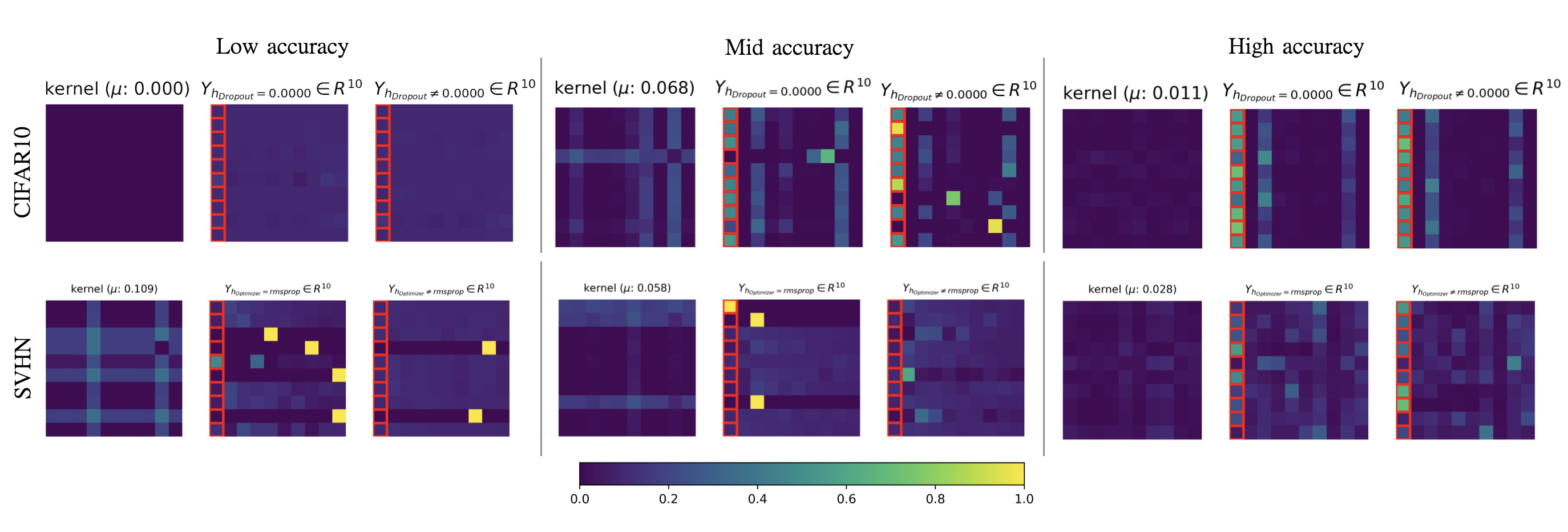}
    \caption{
    Examples of class predictions ($Y_{h=n}(x)$ and $Y_{h \neq n}(x)$) and their dissimilarities ($\left\|\phi(Y_{h=n}(x)) - \phi(Y_{h \neq n}(x))\right\|^2_{\mathcal{G}}$) for different accuracy buckets for CIFAR10 (top) and SVHN (bottom).
    Each row shows 10 random predictions from 3 models in the low- (left), mid- (center), and top- (right) performance buckets, under two different treatment groups for the dropout value ($= 0$ and $\not= 0$). In each performance bucket, there are three subplots. Each subplot is showing 10 randomly selected samples (each row) and their post-softmax values for one of the 10 classes (hence a $10\times10$ grid). The first plot in each trio shows the RBF kernel evaluation of the center and right predictions. The center and right plots show these treatment/control groups. 
    This figure is intended to complement \cref{fig:performance_comparison} to explain why ITE for $Y$ is large for mid-accuracy buckets and small for high-accuracy buckets.
    For CIFAR10, the values are small for low-performing models (most models in this bucket predicting similarly) but for SVHN the values are large due to different diverse predictions.
    %
    }
    \label{fig_app:per_bucket_performance_Y_and_kernel_all}
\end{figure}

\begin{figure}[t]
    \centering
    \includegraphics[width=\textwidth]{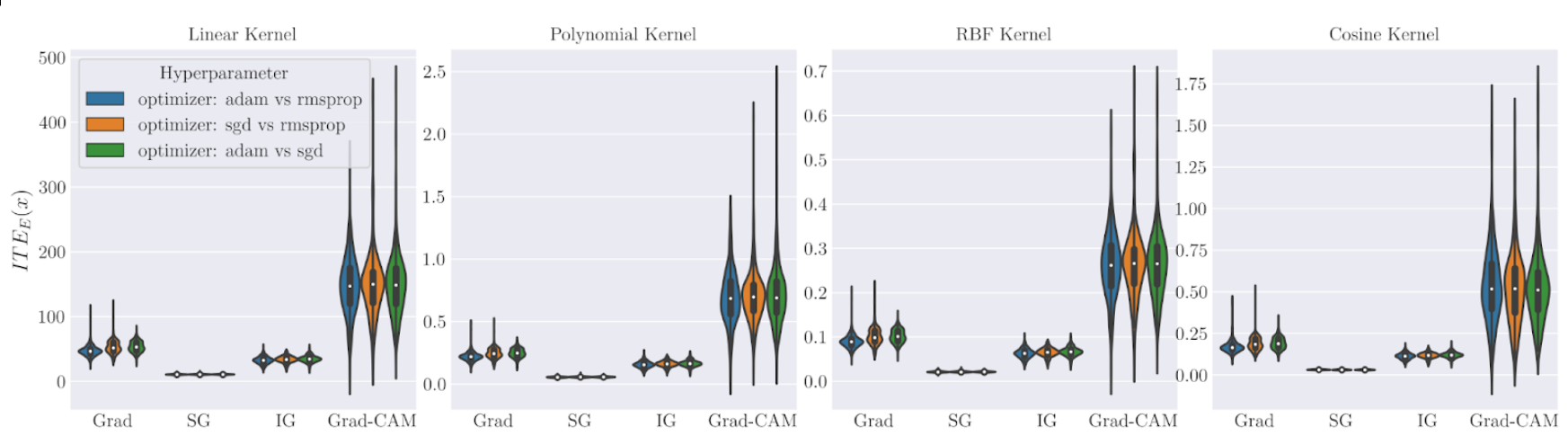} 
    \caption{
        \rebuttal{Comparison of the ITE values with kernelized version of \eqref{eq:extended_te_3} for $E_{h}(x)$ obtained for $100$ instances from CIFAR10 for different choices of the kernel (each column) shows that KTE is not sensitive to the choice of kernels.
        Contrast this figure with \Cref{fig:kernel_comparison}; we conclude that the choice of baseline (i.e., where we contrast \emph{optimizer: adam} against all other optimizers as in \Cref{fig:kernel_comparison} or against other individual values) does not affect the overall trend and should be chosen according to the question in mind: to compare the effect of a hyperparameter value against all other possible values, or against a particular value.}
        %
        %
    }
    \label{fig:kernel_and_baseline_comparison}
\end{figure}

\begin{figure}[h]
    \centering
    \includegraphics[width=\textwidth]{_figures_new/hparam_ite_comparison_cifar10_all_0.000_1.000_rbf.png}
    \includegraphics[width=\textwidth]{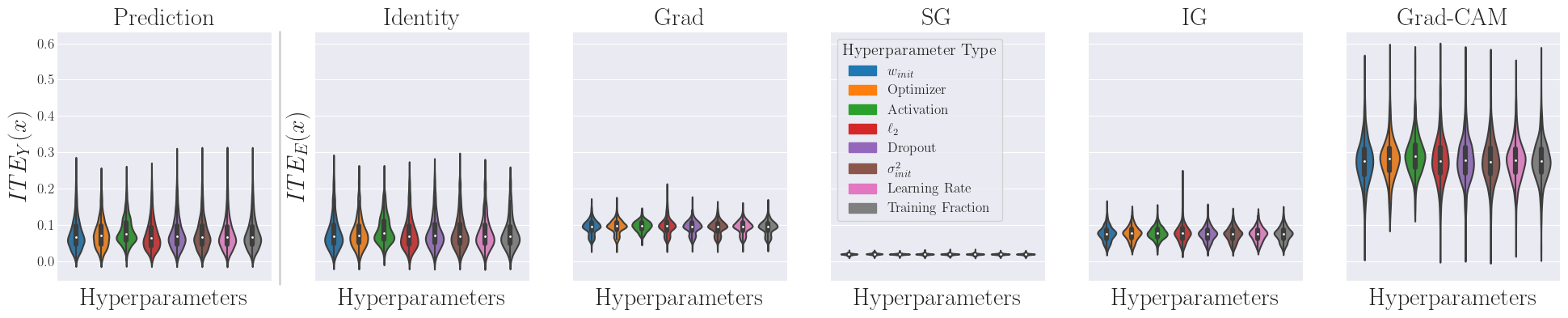}
    \includegraphics[width=\textwidth]{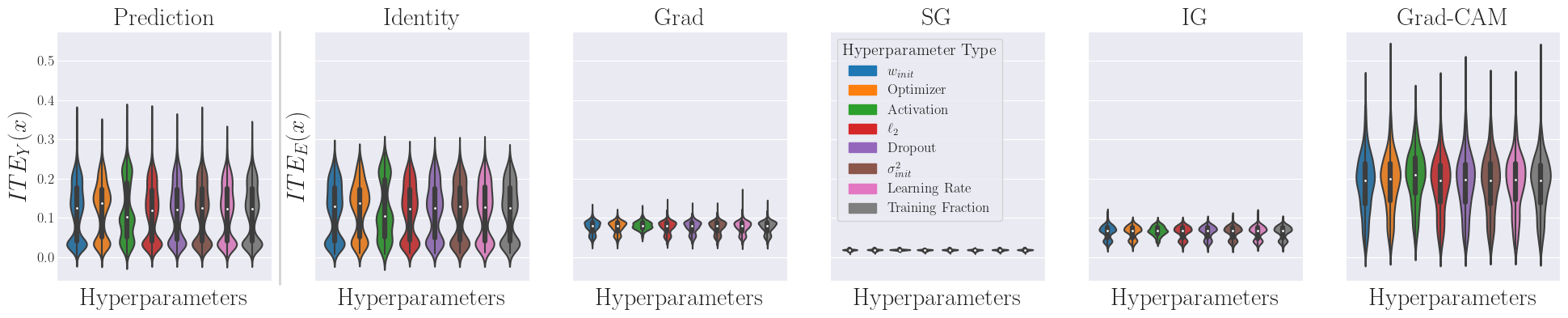}
    \includegraphics[width=\textwidth]{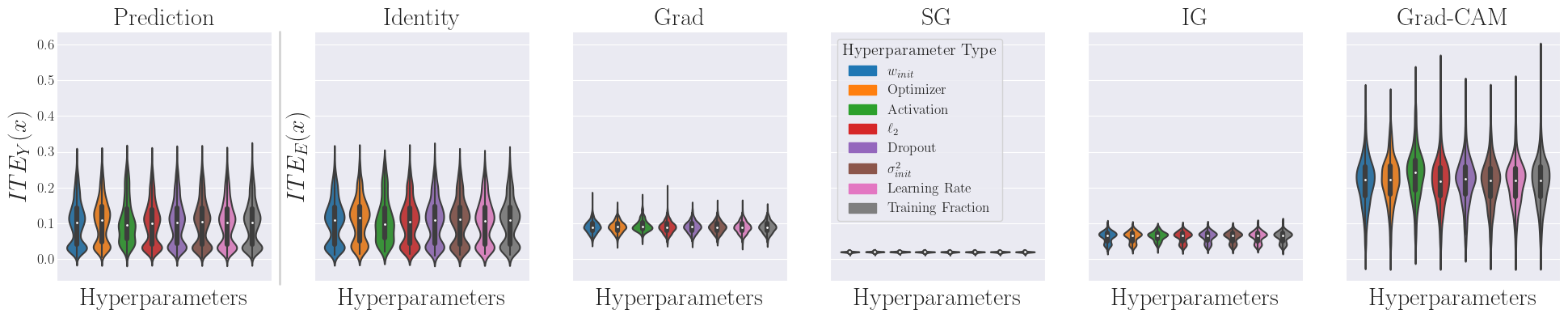}
    \vspace{-2mm}
    \caption{
        ITE values for $Y$ (left) and $E$ (right) show similar effect for different \textit{types} of $H$ across CIFAR10 (row 1), SVHN (row 2), \rebuttal{MNIST (row 3), FASHION (row 4)}.
    }
    \label{fig_app:all_hparam_per_explanation_method}
\end{figure}

\begin{figure}[h]
    \captionsetup[subfigure]{labelformat=empty} 
    \centering
    \includegraphics[width=0.9\textwidth]{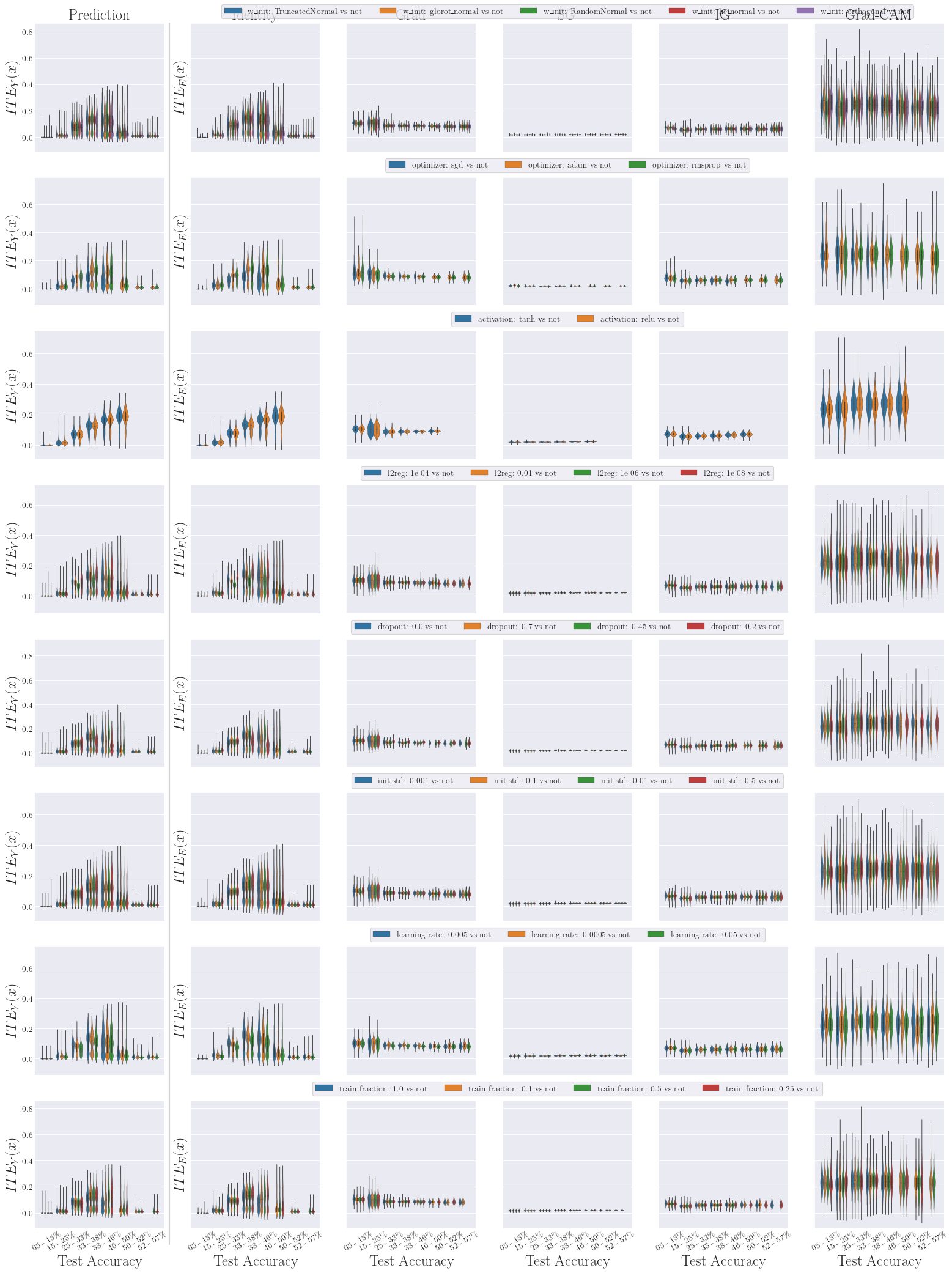}
    \caption{
        Comparison of ITE values of all hyperparameters (each row)on $Y$ (left) and $E$ (right) for models trained on CIFAR10 across different performance buckets, showing the discrepancy in the effect of $H$ on $Y$ vs. that on $E$.
    }
    \label{fig_app:bucket_ite_comparison_cifar10}
\end{figure}

\begin{figure}[h]
    \captionsetup[subfigure]{labelformat=empty} 
    \centering
    \includegraphics[width=0.9\textwidth]{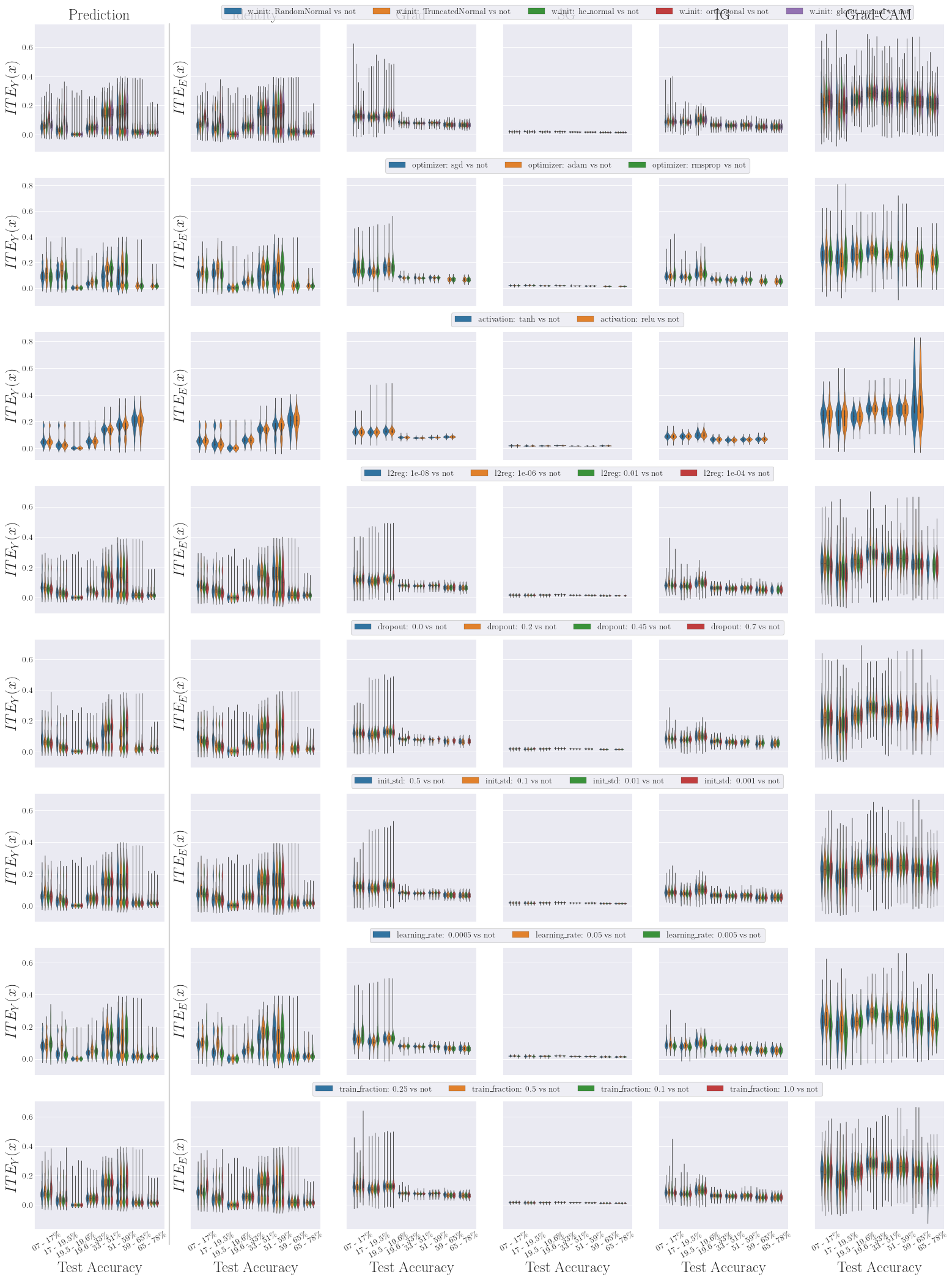}
    \caption{
        Comparison of ITE values of all hyperparameters (each row)on $Y$ (left) and $E$ (right) for models trained on SVHN across different performance buckets, showing the discrepancy in the effect of $H$ on $Y$ vs. that on $E$.
    }
    \label{fig_app:bucket_ite_comparison_svhn_cropped}
\end{figure}

\begin{figure}[h]
    \captionsetup[subfigure]{labelformat=empty} 
    \centering
    \includegraphics[width=0.9\textwidth]{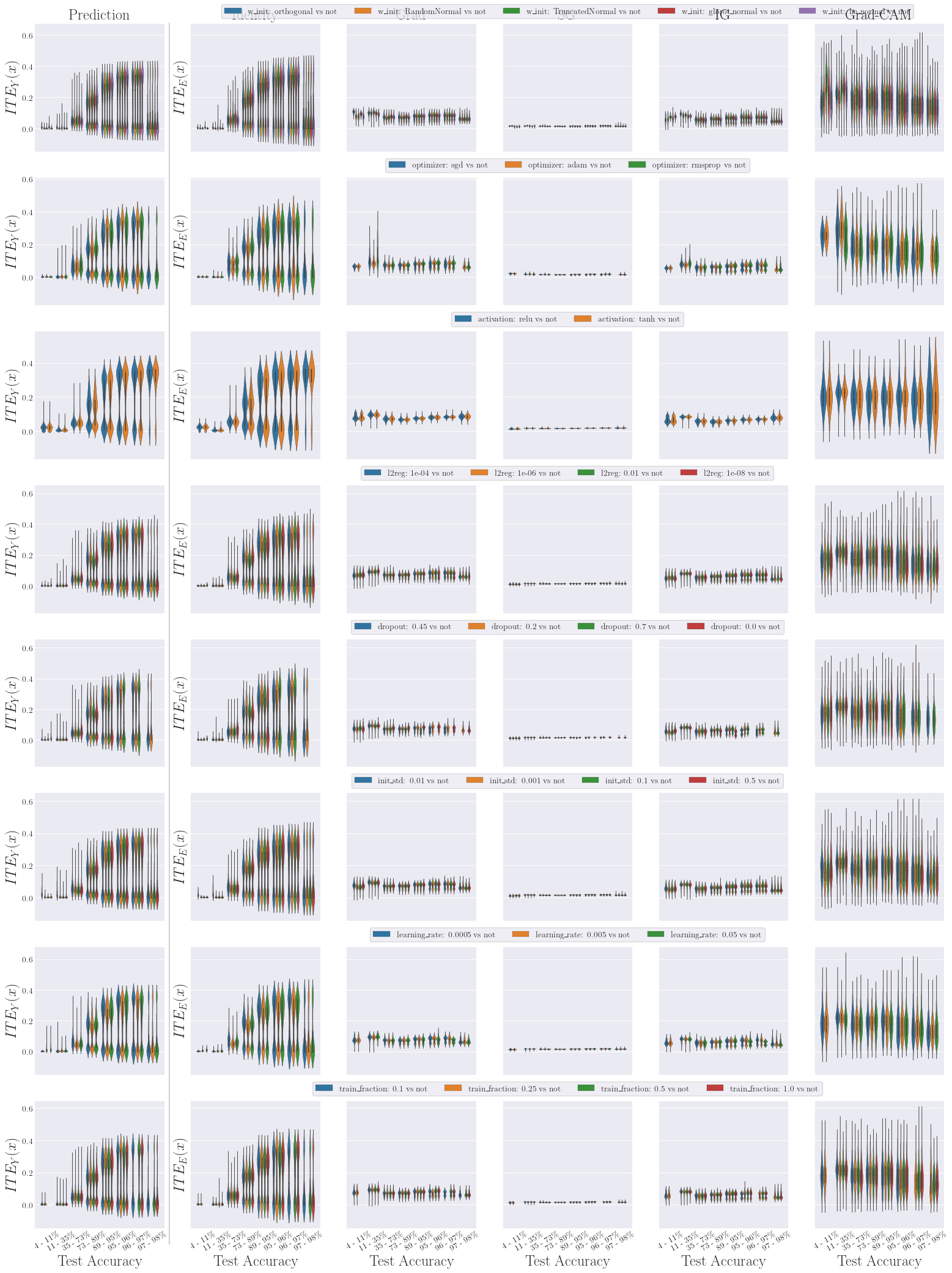}
    \caption{
        \rebuttal{Comparison of ITE values of all hyperparameters (each row)on $Y$ (left) and $E$ (right) for models trained on MNIST across different performance buckets, showing the discrepancy in the effect of $H$ on $Y$ vs. that on $E$.}
    }
    \label{fig_app:bucket_ite_comparison_mnist}
\end{figure}

\begin{figure}[h]
    \captionsetup[subfigure]{labelformat=empty} 
    \centering
    \includegraphics[width=0.9\textwidth]{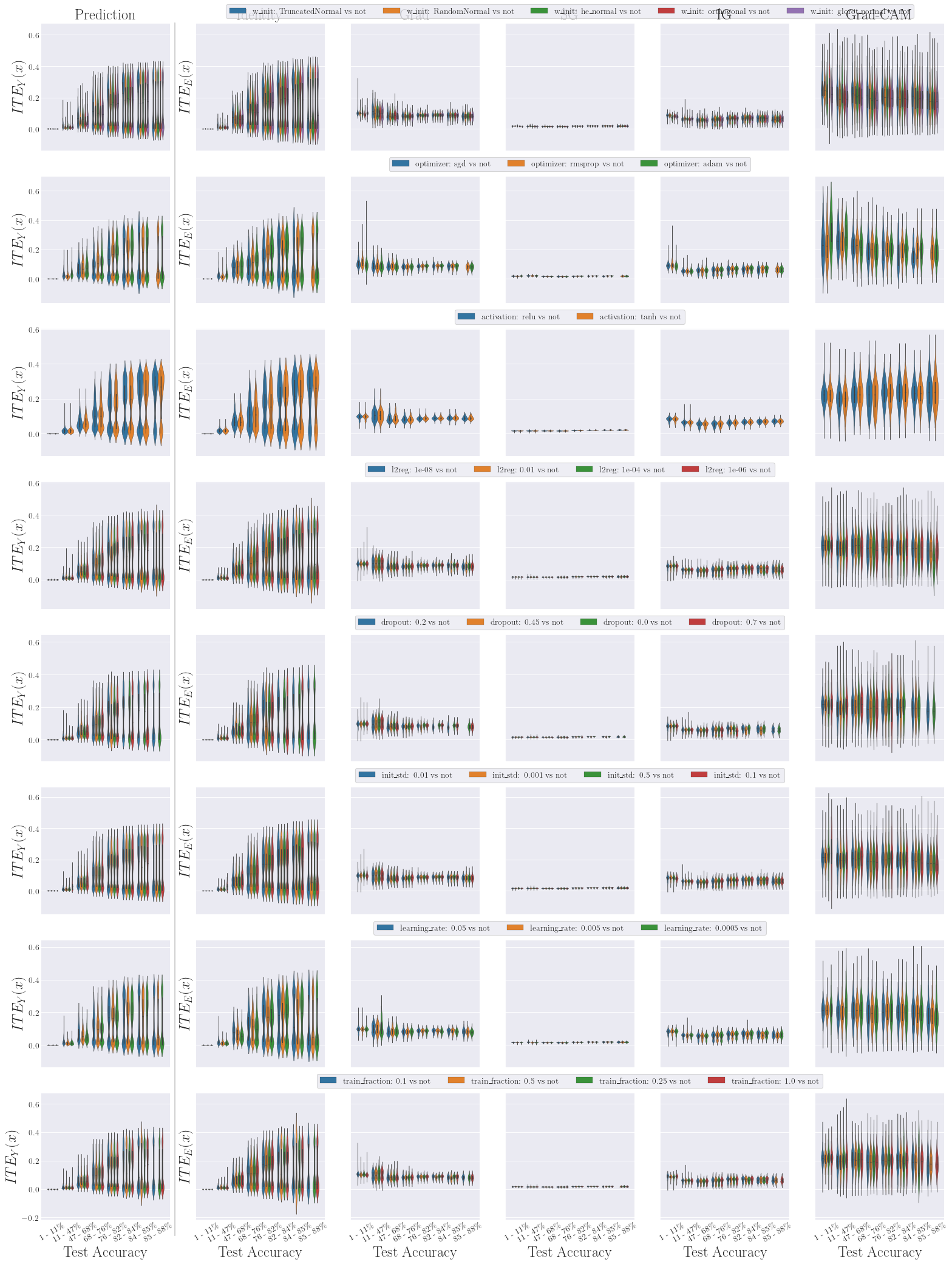}
    \caption{
        \rebuttal{Comparison of ITE values of all hyperparameters (each row)on $Y$ (left) and $E$ (right) for models trained on FASHION across different performance buckets, showing the discrepancy in the effect of $H$ on $Y$ vs. that on $E$.}
    }
    \label{fig_app:bucket_ite_comparison_fashion_mnist}
\end{figure}

\begin{figure}[h]
    \centering
    \includegraphics[width=.8\textwidth]{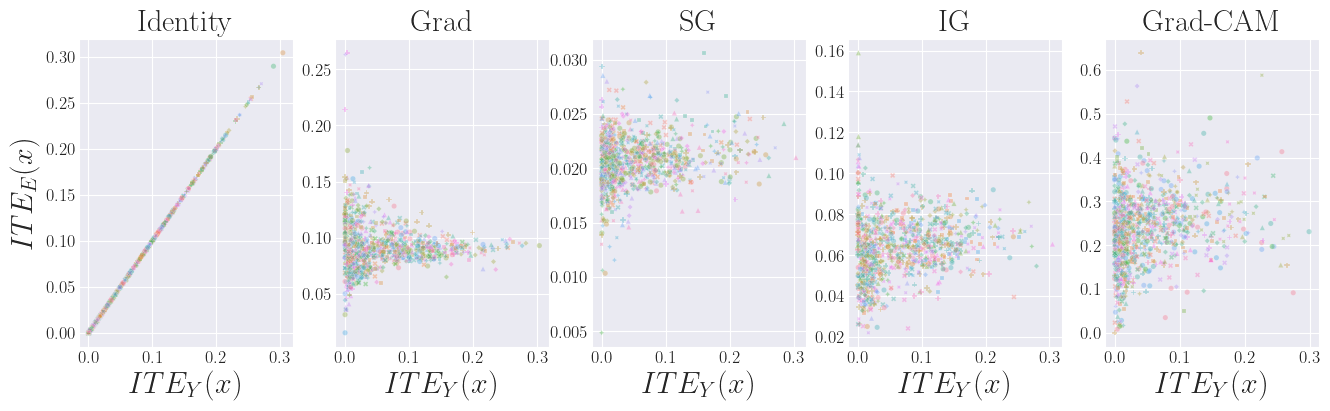}
    \includegraphics[width=.8\textwidth]{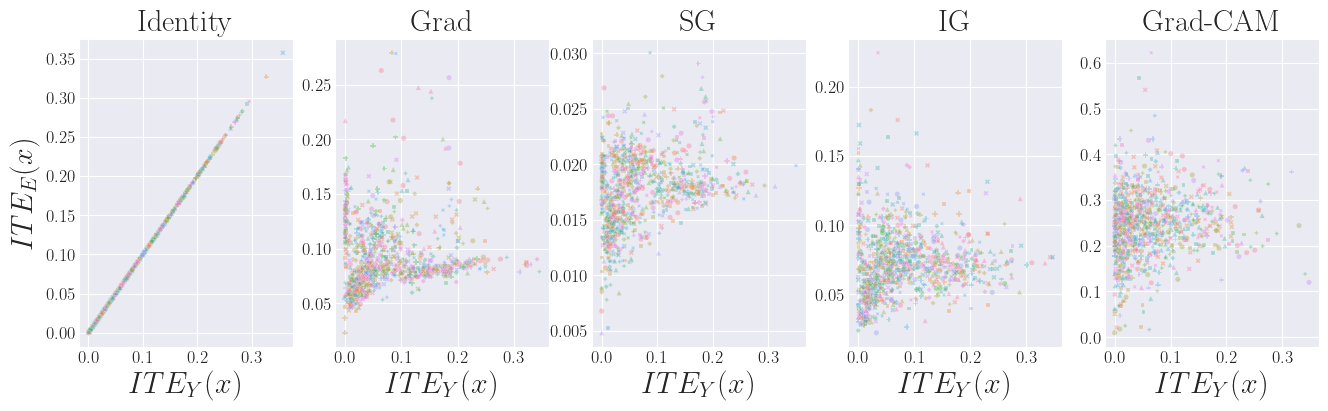}
    \includegraphics[width=.8\textwidth]{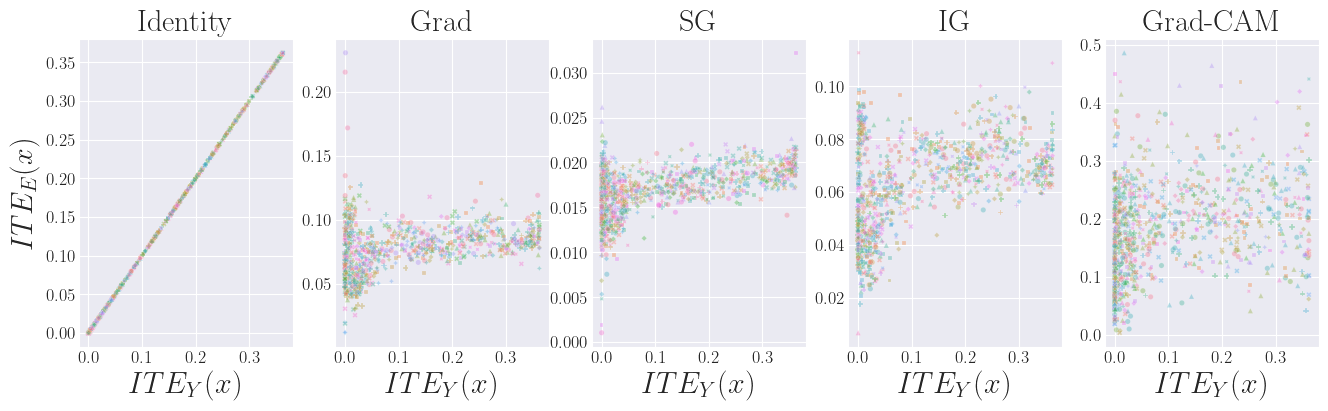}
    \includegraphics[width=.8\textwidth]{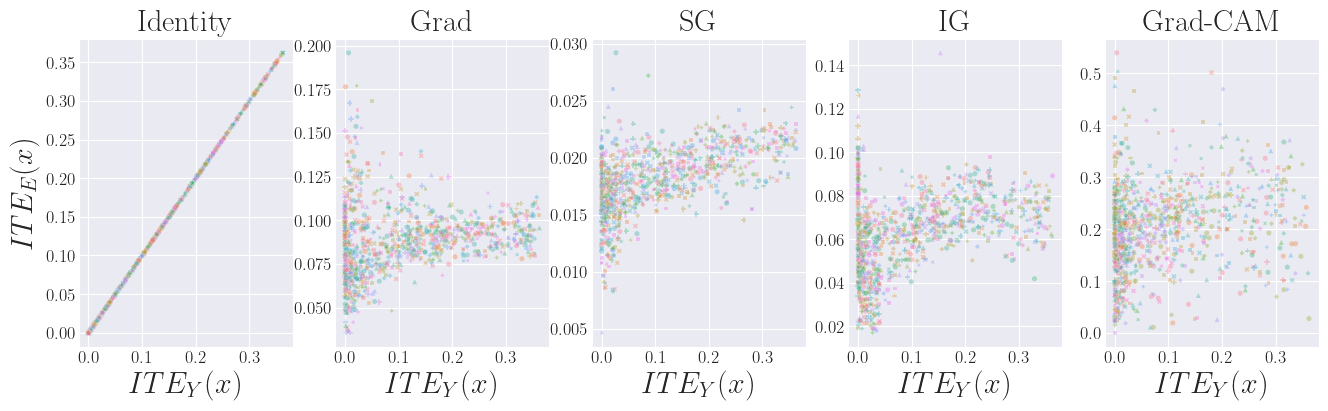}
    \caption{
        Scatter plot of ITE values for $Y$ and $E$ (row 1: CIFAR10; row 2: SVHN; \rebuttal{row 3: MNIST; row 4: FASHION}) across explanation methods reveals no apparent patterns.
    }
    \label{fig_app:scatter_all_hparam_per_explanation_method}
\end{figure}

\begin{figure}[h]
    \centering
    \includegraphics[width=.5\textwidth]{_figures_new/y_vs_e_ite_scattering_cifar10_rbf_all.png} \\\vspace{1em}
    \includegraphics[width=.5\textwidth]{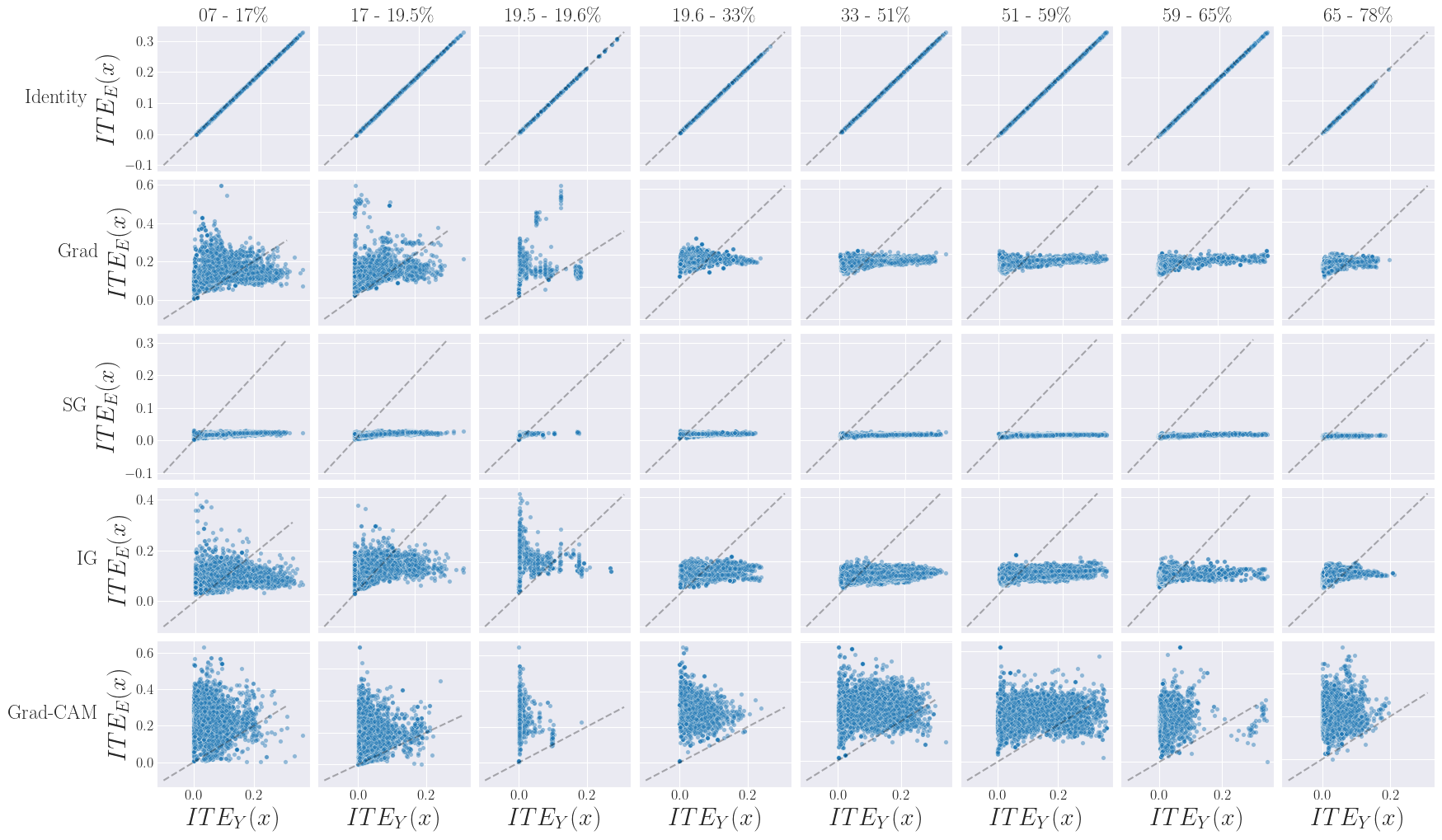} \\\vspace{1em}
    \includegraphics[width=.5\textwidth]{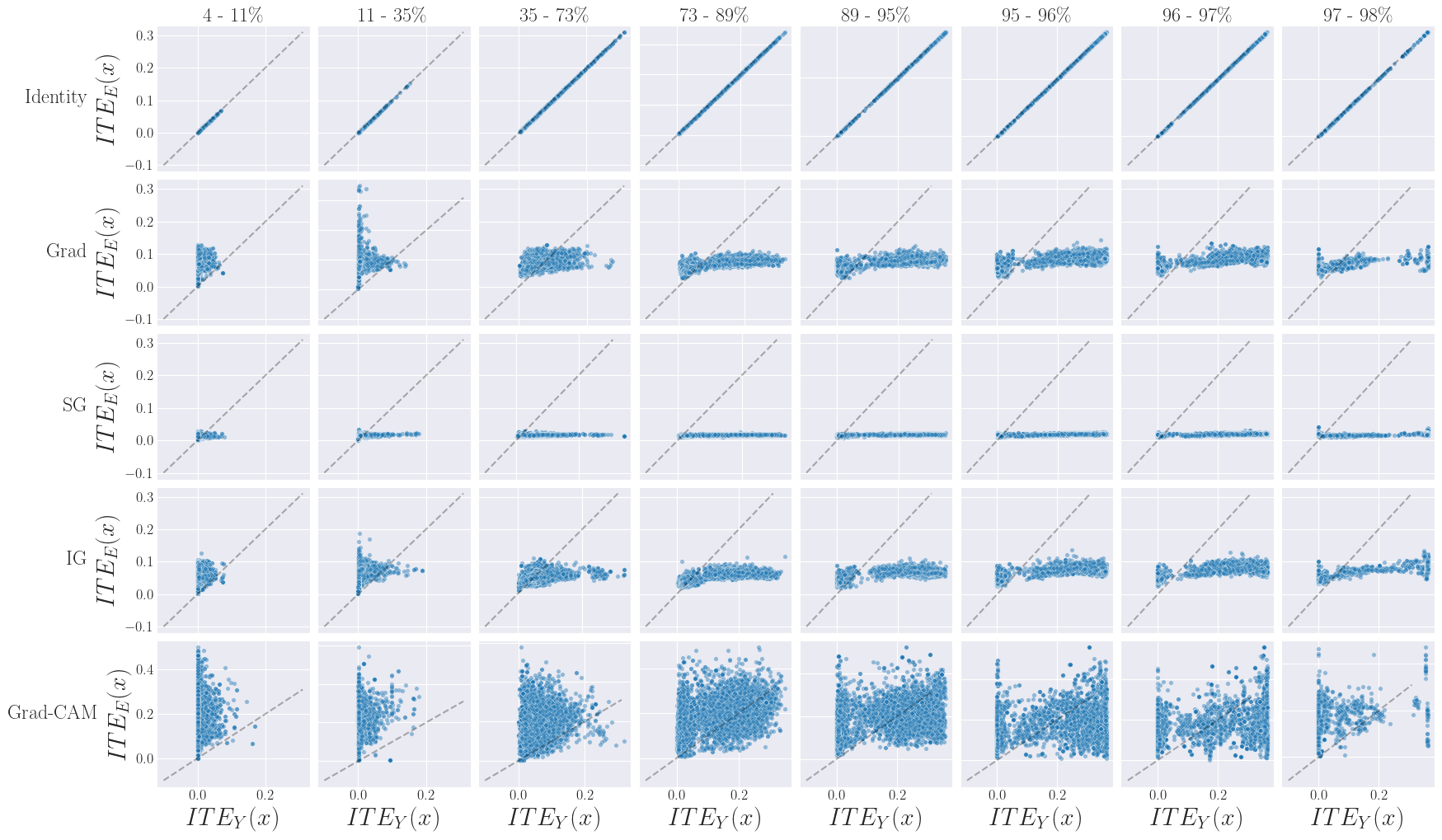} \\\vspace{1em}
    \includegraphics[width=.5\textwidth]{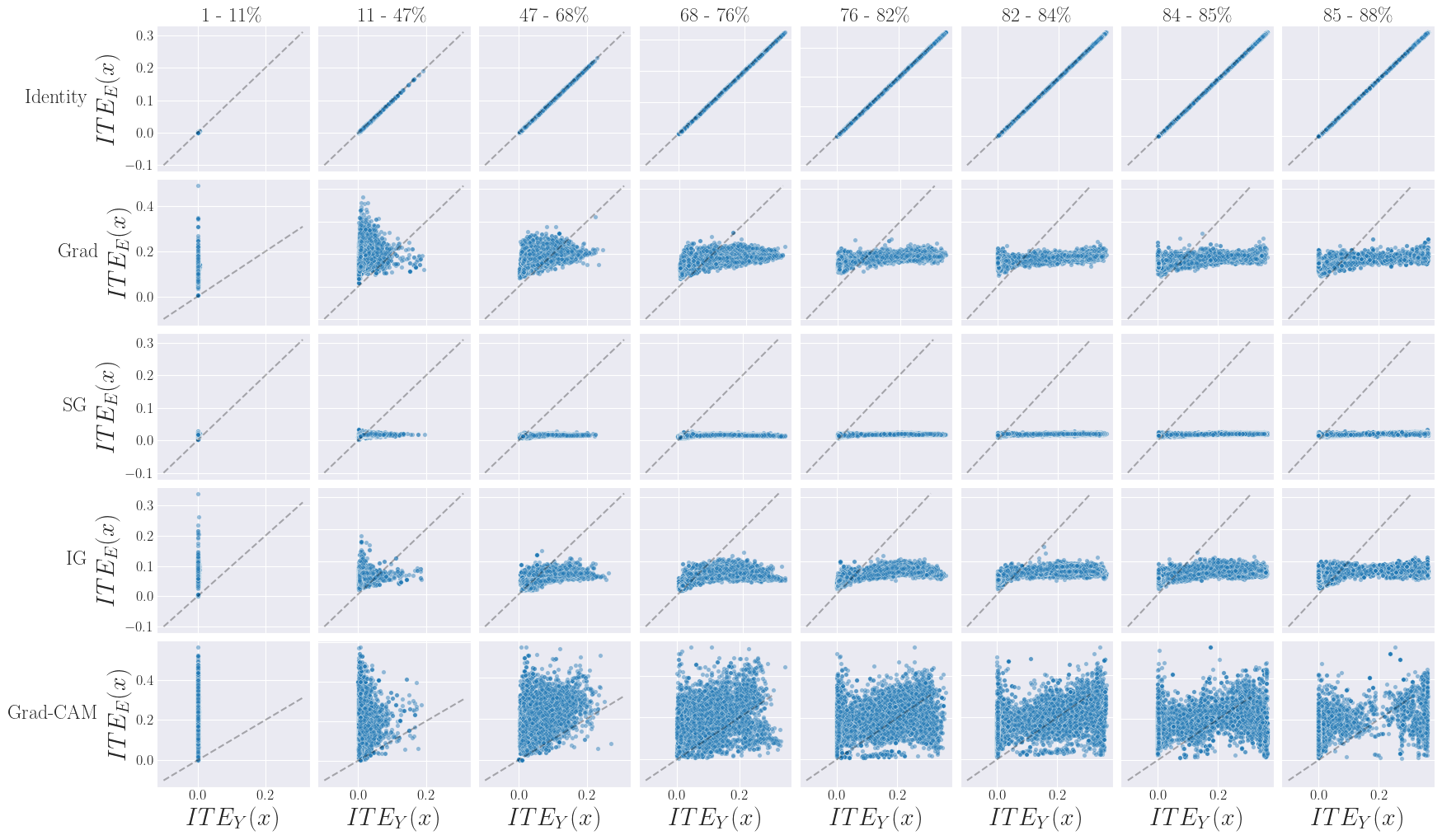}
    \caption{
        Each column is a subset of models at each accuracy bucket, each row is different explanation methods (row 1: CIFAR10; row 2: \rebuttal{SVHN; row 3: MNIST; row 4: FASHION}).
        Whereas low-performing models (first column) show little change in predictions as their explanations differ, top-performing models show the reverse of this trend.
    }
    \label{fig:scatter_ite_y_vs_e_app}
\end{figure}

\begin{figure}[h]
    \centering
    \includegraphics[width=\textwidth]{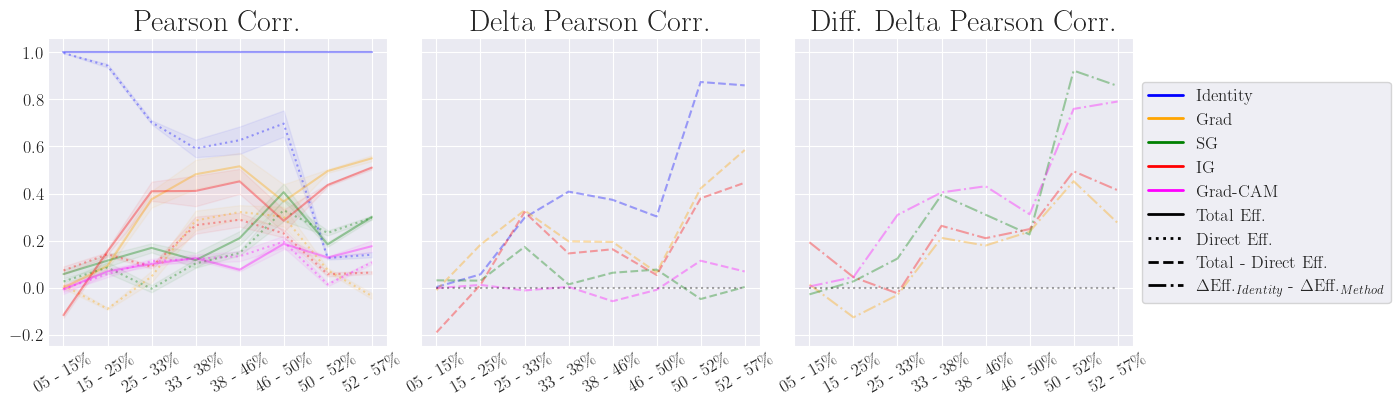}
    \includegraphics[width=\textwidth]{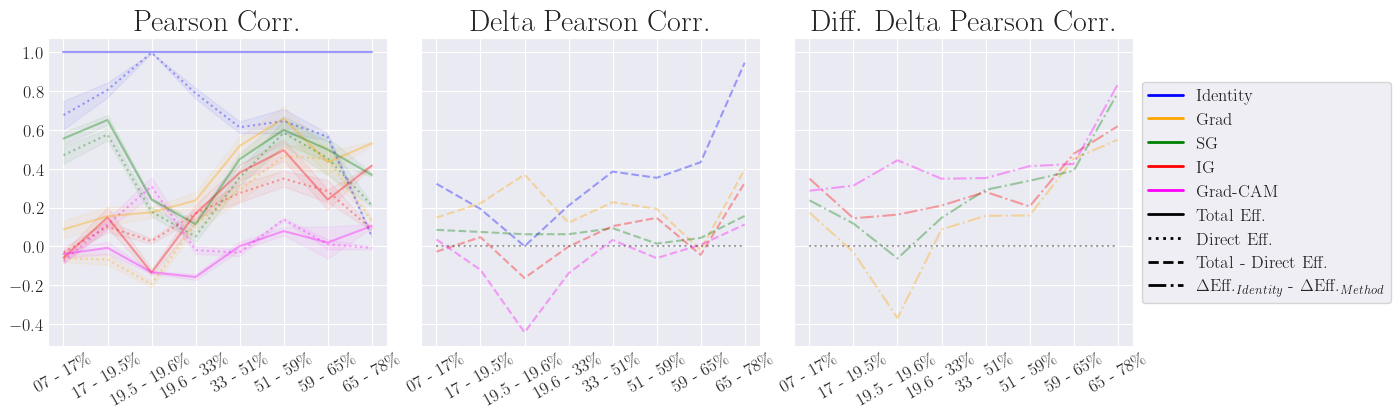}
    \includegraphics[width=\textwidth]{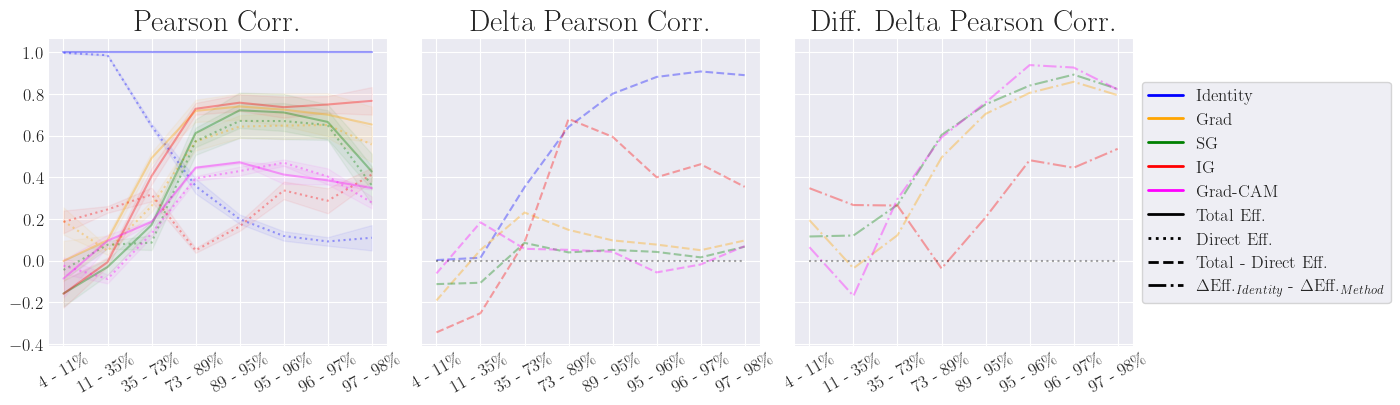}
    \includegraphics[width=\textwidth]{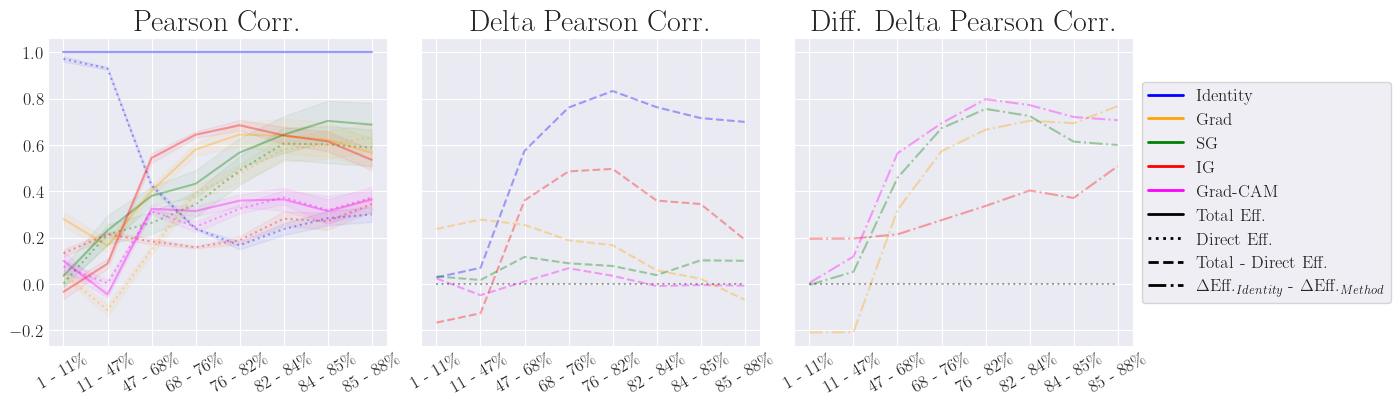}
    \caption{Pearson correlation and Spearman's Rank correlation for ITE of $Y$ and ITE of $E$ across different explanation methods and model performance buckets, for mediated and unmediated $Y$ (row 1: CIFAR10; row 2: SVHN; \rebuttal{row 3: MNIST; row 4: FASHION}). Absolute values of correlation values are smaller across both datasets (max around 0.5), suggesting that $E$ takes influence from $H$ that does not necessarily pass through $Y$. The final absolute correlation is going down for top-performing models in both datasets. The increase in delta correlation between mediated and unmediated $Y$ suggests that the direct impact of $Y$ on $E$ is becoming even more important in top-performing models, even more so for SVHN than for CIFAR10.
    }
    \label{fig:correlation_ite_y_vs_e_app}
\end{figure}

\end{document}